\documentclass[10pt,twocolumn,letterpaper]{article}

\usepackage[pagenumbers]{cvpr} % To force page numbers, e.g. for an arXiv version

\usepackage{booktabs}
\usepackage[table]{xcolor}
\usepackage{booktabs}
\usepackage{graphicx}
\usepackage[table]{xcolor} % \rowcolor 를 위해
\usepackage{cuted}   % \begin{strip} ... \end{strip}
\usepackage{capt-of} % \captionof{figure}{...}
\usepackage{multirow}
\usepackage{array}
\usepackage{makecell}
\usepackage{calc}

\definecolor{rowgray}{gray}{0.92} % Ours 행에 쓸 연한 회색
\definecolor{lightgray}{gray}{0.9}

\definecolor{darkred}{rgb}{0.55, 0.0, 0.0}

\definecolor{cvprblue}{rgb}{0.21,0.49,0.74}
\usepackage[pagebackref,breaklinks,colorlinks,allcolors=cvprblue]{hyperref}

\newcommand{\myparagraph}[1]{\vspace{2pt}\noindent{\bf #1}}

\def\paperID{} % *** Enter the Paper ID here
\def\confName{CVPR}
\def\confYear{2026}

\title{WaTeRFlow: Watermark Temporal Robustness via Flow Consistency}

\author{%
Utae Jeong$^{1}$ \quad
Sumin In$^{1}$ \quad
Hyunju Ryu$^{1}$ \quad
Jaewan Choi$^{1}$\\
Feng Yang$^{2}$ \quad
Jongheon Jeong$^{1}$ \quad
Seungryong Kim$^{3}$ \quad
Sangpil Kim$^{1}$\thanks{Corresponding Author}\\[0.5em]
$^{1}$ Korea University \quad
$^{2}$ Google DeepMind \quad
$^{3}$ KAIST AI
}

\begin{document}
\maketitle
\begin{strip}
  \centering
  % \includegraphics[width=\textwidth]{figure/figure1_06.pdf}
  % Tighten vertical space so Abstract fits entirely in the left column
  \setlength{\abovecaptionskip}{2pt}
  \setlength{\belowcaptionskip}{-2pt}
  \includegraphics[width=0.98\textwidth]{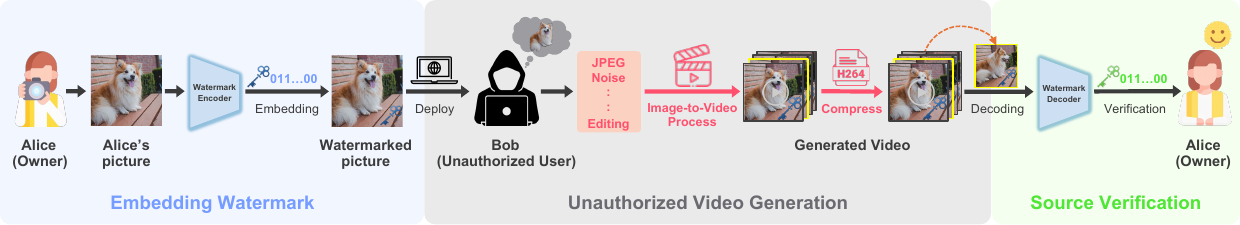}
  \captionof{figure}{\textbf{Application scenario of WaTeRFlow.} Specifically, we consider watermark embedding for image copyright protection and verification from generated videos. Left: An owner, Alice, protects her picture by embedding a watermark. Middle: An unauthorized user, Bob, generates a video from Alice’s picture. Right: On Alice’s side, copyright is verified by recovering the watermark from the video frames using her watermark decoder, in order to identify the source used to generate the video and verify whether her picture was used.}
  \label{fig1}
  \vspace{-0.2em}
\end{strip}
\begin{abstract}
Image watermarking supports authenticity and provenance, yet many schemes are still easy to bypass with various distortions and powerful generative edits. Deep learning-based watermarking has improved robustness to diffusion-based image editing, but a gap remains when a watermarked image is converted to video by image-to-video (I2V), in which per-frame watermark detection weakens. I2V has quickly advanced from short, jittery clips to multi-second, temporally coherent scenes, and it now serves not only content creation but also world-modeling and simulation workflows, making cross-modal watermark recovery crucial. We present \textbf{WaTeRFlow}, a framework tailored for robustness under I2V. It consists of (i) FUSE (Flow-guided Unified Synthesis Engine), which exposes the encoder–decoder to realistic distortions via instruction-driven edits and a fast video diffusion proxy during training, (ii) optical-flow warping with a Temporal Consistency Loss (TCL) that stabilizes per-frame predictions, and (iii) a semantic preservation loss that maintains the conditioning signal. Experiments across representative I2V models show accurate watermark recovery from frames, with higher first-frame and per-frame bit accuracy and resilience when various distortions are applied before or after video generation.
\end{abstract}    
\vspace{-20pt}
\section{Introduction}
\label{sec:intro}
Diffusion models~\cite{diffusion} have made image editing, including text-guided edits~\cite{instructpix2pix, ultra_edit, prompt_to_prompt}, highly reliable. Digital watermarking embeds a message into an image to support authenticity and provenance while remaining imperceptible. Robustness requires accurate recovery after common transforms and edits. Recent watermarking methods are designed for robustness, and embedded watermarks often remain detectable even after editing with diffusion models. Unlike typical image distortions such as JPEG compression that perturb a fixed canvas, image-to-video (I2V) synthesizes temporally coherent frames from a single image, which often acts as a watermark-disrupting transformation by weakening the watermark signal across frequencies and inducing temporally varying subpixel misalignments that make per-frame recovery intrinsically difficult~\cite{vine}.

We study a provenance scenario shown in Fig.~\ref{fig1}. Alice embeds a watermark in her picture. Bob, an unauthorized user, inputs that image to an I2V model and produces a video. He may further apply edits, regeneration, compression, or other manipulations before or after video synthesis to suppress the watermark signal. Using the generated video solely, Alice must recover the watermark with high average accuracy and consistently high per-frame accuracy across the frames to verify ownership of the source image.

To address this challenge, we introduce WaTeRFlow, a watermarking and detection framework for I2V. Our key idea is to couple realistic video-induced effects with warping and temporal regularization when learning the encoder and decoder. We operationalize this via the Flow-guided Unified Synthesis Engine (FUSE), inserted between the encoder and decoder. FUSE unifies diffusion-based image edits and I2V within a single loop by integrating an instruction-driven image editing proxy~\cite{instructpix2pix} and a fast video diffusion proxy~\cite{animatelcm}, exposing the system to realistic pixel transformations with low overhead. We warp each generated frame to the first frame using an optical-flow estimator~\cite{raft} to control the strength of distortions induced by video generation seen by the encoder and decoder, thereby stabilizing learning. We feed edited images and flow-warped frames to the decoder. We apply a semantic preservation loss to keep the conditioning signal used by the video generator close to that of the original image while enforcing a Temporal Consistency Loss (TCL) on adjacent flow-warped frames to stabilize per-frame predictions.

We evaluate the transferability of robustness from video diffusion proxy to two representative I2V models. Specifically, we test on the U-Net based Stable Video Diffusion (SVD)~\cite{stablevideodiffusion} and the Diffusion Transformer (DiT) based CogVideoX~\cite{cogvideox}, yielding consistent gains on both. WaTeRFlow improves both first-frame and per-frame bit accuracy while mitigating late-frame degradation in watermark detection. These outcomes are crucial for verifying content origin in real-world scenarios. For example, on SVD-XT the average bit accuracy increases from 73.92\% with VINE~\cite{vine} to 84.96\%, and the first frame reaches 96.93\%.

\noindent Our main contributions are summarized as follows:
\begin{itemize}
\item We optimize the system end-to-end with FUSE, which inserts image editing proxy and video diffusion proxy between the encoder and decoder. In addition, we perform flow-guided warping of each generated frame to the first frame, enabling watermark recovery even when manipulations occur before or after video synthesis.
\item We introduce a semantic preservation loss during watermark embedding to retain conditioning derived from the watermarked image that guides I2V generation, thereby improving first-frame accuracy.
\item We apply a TCL to adjacent flow-warped frames, stabilizing frame-wise outputs and reducing variance over time.
\end{itemize}
\section{Related Work}
\label{sec:rel}
\myparagraph{Steganography and Digital Watermarking.} The rapid growth and easy distribution of digital media challenge both information confidentiality and intellectual property rights. Information hiding technologies evolved to address these needs, branching into steganography and digital watermarking, which differ in their primary objectives. Steganography~\cite{stegastamp, hinet} prioritizes invisibility, covertly embedding information within digital content. In contrast, digital watermarking protects assets by embedding rights-identifying data, prioritizing robustness against various distortions and attacks. Early digital watermarking research focused on the spatial~\cite{digital, watermark} and frequency domains~\cite{wavelet_wm, dwt_wm}. Subsequently, deep learning-based methods~\cite{distortion, hidden} jointly optimize invisibility and detection stability. These advancements have expanded watermarking applications from 2D images~\cite{stable_signature, cin, tree_ring,robin} to videos~\cite{lvmark, videoshield,rivagan,videoseal}. 
% This study focuses on image watermarking and shows existing methods are fragile to image-to-video (I2V) generation, which motivates our design choices.

\myparagraph{Robust Image Watermarking.} Recent watermarking research has rapidly expanded, focusing on robustness against the diverse distortions and transformations in real-world distribution and editing environments. HiDDeN~\cite{hidden} proposed an approach that jointly learns watermarking and robustness by inserting a layer that simulates distortions such as Gaussian blur and JPEG. TrustMark~\cite{trustmark} introduced a method robust to resolution changes by employing resolution scaling, while WAM~\cite{wam} proposed one robust to partial edits through pixel-level detection and decoding. Following the advancement of diffusion-based generative models and reports that variational autoencoder (VAE)~\cite{vae} and diffusion-based editing can easily remove watermarks~\cite{watermark_ai_removal_survey}, Robust-Wide~\cite{robustwide} targeted instruction-driven image editing~\cite{instructpix2pix, magic_brush, ddim_inversion}. VINE~\cite{vine} likewise addressed editing and proposed a benchmark integrating various distortions, including an I2V pipeline~\cite{stablevideodiffusion}. However, robustness to I2V remains a significant challenge for these methods. 
% In this paper, we propose a new watermarking technique specifically designed for I2V robustness.

\myparagraph{Video Generation Models.} With the recent advances in diffusion-based image generation models~\cite{image_diffusion, ddpm, latent_image_diffusion}, video generation models have evolved by adopting U-Net~\cite{unet} or the Diffusion Transformers (DiT)~\cite{dit} as their backbones. The video diffusion model (VDM)~\cite{video_diffusion} extended diffusion models to the video domain by modifying the 2D U-Net from image diffusion models into a spatial-temporal network. To reduce computational costs, Latent Video Diffusion Models (LVDMs)~\cite{latent_video_diffusion} compress videos into latent features. Stable Video Diffusion (SVD)~\cite{stablevideodiffusion} is a latent video diffusion model with a U-Net backbone, formalizing the video generation process by extending image diffusion temporally. In contrast, CogVideoX~\cite{cogvideox} leverages a DiT backbone with a 3D VAE and an expert transformer to enhance text-video alignment~\cite{videobooth} and generate high-quality video. Despite these advancements, diffusion-based video generation models suffer from limitations in generation speed and memory efficiency due to their structural reliance on multi-step sampling. To address this, various methods using consistency models~\cite{consistency} have been proposed. Latent Consistency Model (LCM)~\cite{lcm} effectively improves memory usage and inference speed via consistency learning in the latent space, and AnimateLCM~\cite{animatelcm} extended this to the video domain, achieving fast sampling while maintaining strong temporal consistency and visual quality.
\section{Preliminaries}
\label{sec:pre}
\myparagraph{Stable Video Diffusion
(SVD)}~\cite{stablevideodiffusion} performs image-to-video (I2V) generation in latent space as follows:
\begin{equation}
z_{\text{img}}=\text{enc}(I), \quad
\mathbf{Z}^{(0)} \sim \mathcal{N}\big( \mathbf{0}, \sigma_{\max}^2 \mathbf{I} \big),
\label{eq:svd00}
\end{equation}
\noindent where $\text{enc}(\cdot)$ denotes the variational autoencoder (VAE) encoder, $z_{\text{img}}\in \mathbb{R}^{C_{\text{lat}} \times H_{\text{lat}} \times W_{\text{lat}}}$ is the encoded image latent representation, and $\mathbf{Z}^{(0)} \in \mathbb{R}^{T \times C_{\text{lat}} \times H_{\text{lat}} \times W_{\text{lat}}}$ is the initial latent noise. Here, $T$ is the number of frames, $C_{\text{lat}}$ is the number of latent channels, and $H_{\text{lat}}$, $W_{\text{lat}}$ are the downsampled spatial sizes in latent space. Let $c$ denote the conditioning signal derived from the input image and optional metadata (e.g., $z_{\text{img}}$, CLIP~\cite{clip} image embedding, and frame rate). Using a finite-step sampler with a decreasing noise schedule $\sigma_0=\sigma_{\max} > \sigma_1 > \cdots > \sigma_N \approx 0$, the denoiser $D_\theta$ iteratively updates the latent variables for $j=1,\dots,N$ as:
\begin{equation}
\begin{split}
\mathbf{Z}^{(j)}=f_{\text{sampler}}&\big(\mathbf{Z}^{(j-1)},\\
& D_{\theta}(\mathbf{Z}^{(j-1)},\sigma_{j-1},c),\sigma_{j-1},\sigma_j\big),
\label{eq:svd01}
\end{split}
\end{equation}
\noindent where $f_{\text{sampler}}$ denotes the one-step update rule of the chosen sampler (e.g., Euler), which maps the current latent, the denoiser output at $\sigma_{j-1}$, and the pair \((\sigma_{j-1}, \sigma_{j})\) to the next latent. Finally, the latent variable at the last step $\mathbf{Z}^{(N)}$ is decoded to obtain the video as:
\begin{equation}
V = \{v_t\}_{t=0}^{T-1}=\text{dec}\big(\mathbf{Z}^{(N)}\big),
\label{eq:svd02}
\end{equation}
\noindent where $\text{dec}(\cdot)$ denotes the VAE decoder that projects the final latent variable $\mathbf{Z}^{(N)}$ back to pixel space.

\myparagraph{AnimateLCM}
~\cite{animatelcm} adopts the same latent initialization and decoding as SVD in Eq.~\ref{eq:svd00} and Eq.~\ref{eq:svd02}. The distinction lies in the denoising transition of Eq.~\ref{eq:svd01}, where the diffusion denoiser $D_\theta$ is replaced by a consistency-distilled denoiser $C_\phi$, and the sampler in Eq.~\ref{eq:svd01} uses only $K\!\ll\!N$ steps. Without classifier-free guidance (CFG)~\cite{cfg}, each step is a single forward pass, yielding substantial speed and memory savings at high output quality. Style and acceleration are injected as lightweight residuals into a base weight $\eta_0$:
\begin{equation}
\eta' = \eta_0 + \alpha\,\tau_{\text{personalized}} + \beta\,\tau^{\text{image}}_{\text{acc}} + \gamma\,\tau^{\text{video}}_{\text{acc}},
\label{eq:animlcm_weights}
\end{equation}
\noindent where $\tau_{\text{personalized}}$ is a LoRA residual~\cite{lora}, and $\tau^{\text{image}}_{\text{acc}}$ and $\tau^{\text{video}}_{\text{acc}}$ are acceleration residuals. The residuals are merged into $\eta'$ without extra inference FLOPs.
\begin{figure*}
\centering
\includegraphics[width=\textwidth]{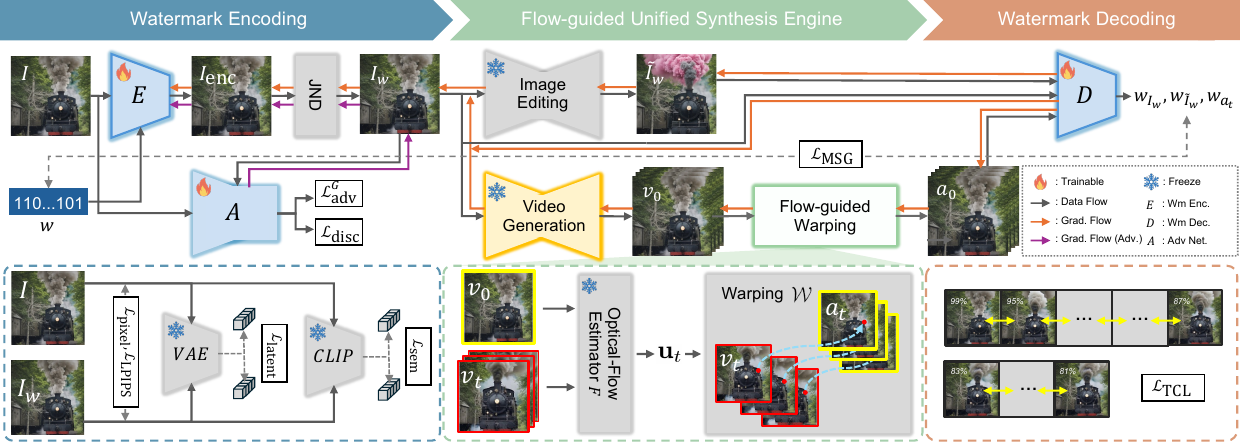}
\vspace{-10pt}
\caption{\textbf{The overview of WaTeRFlow.} Left: The watermark encoder is optimized to embed the watermark while preserving quality in both pixel and latent space, and it is trained to keep the watermarked image semantically close to the original. Middle: Image editing and video generation are performed by an image editing proxy and a video diffusion proxy, respectively, and the generated frames are then warped to the first frame. Right: The decoder processes the images produced by FUSE to decode the embedded watermark and compute the training loss. Overall, these components enable watermark insertion and detection that are robust to image-to-video generation.}
\label{fig2}
\vspace{-10pt}
\end{figure*}
\section{Method}
\label{sec:method}
We aim to recover a $k$-bit watermark with high accuracy from frames synthesized using a single watermarked image. Let $I \in \mathbb{R}^{C \times H \times W}$ denote the original image and $w \in \{0,1\}^k$ the watermark message. The encoder $E$ generates the watermarked image $I_w=E(I,w)$. An image-to-video (I2V) generator, such as Stable Video Diffusion (SVD)~\cite{stablevideodiffusion}, conditions on $I_w$ and produces frames $V=\{v_t\}_{t=0}^{T-1}$. The decoder $D$ predicts the watermark from each frame $v_t$. As shown in Fig.~\ref{fig2}, our methodology comprises three components: watermark encoding, a module for enhancing watermark robustness, and watermark decoding. We present them in that order for clarity and coherence.

\subsection{Watermarking Encoder}
The encoder uses a message-conditioned U-Net. We first interpret the message bits as a square bit grid with side length $m=\sqrt{k}$, where $k$ is the message length. The bit vector is reshaped into a binary grid of size $1\times m \times m$, then lifted to a learned feature map using a Convolutional Neural Network (CNN) and upsampling. This feature map is concatenated with the input image along the channel dimension and passed through a U-Net to produce watermark signal. To preserve perceptual quality, we modulate only the residual between the raw output and the original image. Specifically, we compute a data-dependent scale map from a just-noticeable difference (JND)~\cite{jnd} heatmap. This map has one channel and the same spatial resolution as the image, and it is bounded away from zero to prevent the watermark signal from disappearing. The final watermarked image is obtained by adding the element-wise product of the scale map and the residual back to the input image. This encourages the encoder to concentrate watermark signal in regions where changes are least visible while keeping the embedding aware of the image. We further enhance imperceptibility via adversarial training, as detailed in Sec.~\ref{optimization}.

\myparagraph{Semantic Preservation Loss.}
To keep high-level semantics stable, we regularize the CLIP~\cite{clip} image embeddings of the watermarked image \(I_w\) and the original image \(I\):
\begin{equation}
\mathcal{L}_{\text{sem}}
=1-\cos(f_\text{CLIP}(I_w),f_\text{CLIP}(I)),
\label{eq:cliploss}
\end{equation}
\noindent where $f_{\text{CLIP}}$ denotes the pretrained, frozen CLIP image encoder~\cite{clip} that maps an image $I$ to a unit-norm embedding. As a term in the overall training objective, the semantic preservation loss keeps the watermarked image semantically consistent with the original and maintains the conditioning $c$ used by a diffusion-based I2V model~\cite{stablevideodiffusion} during video generation. With $c$ derived from the condition image, the loss further encourages $c$ to be less affected by watermark signals when the video is generated from the watermarked image rather than from the original. Empirically, enforcing semantic preservation enables the watermark signal to survive the I2V process. In SVD, where the U-Net's attention uses the condition image's CLIP embedding as \textit{keys} and \textit{values}, the first frame $v_0$ remains close to $I_w$, yielding markedly higher bit accuracy of the first frame.

\subsection{FUSE: Flow-guided Unified Synthesis Engine}
\label{subsec:fuse}
\noindent To train for robustness under both instruction-driven edits and I2V, we introduce a module named Flow-guided Unified Synthesis Engine (FUSE) between the watermark encoder and decoder. FUSE integrates an image editing branch and a video generation branch. During training, we optimize the encoder and decoder end-to-end with FUSE.

\myparagraph{Image Editing Branch.}
Following~\cite{robustwide}, we employ InstructPix2Pix~\cite{instructpix2pix} as a proxy model for image editing. During training, this model edits the watermark-embedded image using text prompts, and the watermark decoder takes the edited image as input to recover the watermark.

\myparagraph{Video Generation Branch.}
The video generation branch comprises a video diffusion proxy and an optical- flow estimator that warps the generated frames to a reference frame. We designate the first generated frame $v_0$ as the reference and warp each $v_t$ toward $v_0$ with RAFT~\cite{raft} to mitigate generation-induced distortions and stabilize learning:
\begin{subequations}\label{eq:flow-group} % (6)
\begin{equation}
a_t=\mathcal{W}\big(v_t,\mathbf{u}_t\big), \quad 
\mathbf{u}_t = F(v_0,\,v_t)\in\mathbb{R}^{H\times W\times 2},
\label{eq:flow} % (6a)
\end{equation}
\begin{equation}
[\mathcal{W}(v,\mathbf{u})](\mathbf{p})=v\big(\mathbf{p}+\mathbf{u}(\mathbf{p})\big),
\label{eq:warp-def} % (6b)
\end{equation}
\end{subequations}
\noindent where $F$ is the optical-flow estimator and $\mathbf{u}_t$ is the forward optical-flow field from $v_0$ to $v_t$. $a_t$ denotes the frame warped toward $v_0$ by the warping function $\mathcal{W}$. Here, $\mathcal{W}$ is a bilinear backward warper. The coordinate $\mathbf{p}$ lies on the reference grid of $v_0$, so $[\mathcal{W}(v_t,\mathbf{u}_t)](\mathbf{p})$ samples $v_t$ at $\mathbf{p}+\mathbf{u}_t(\mathbf{p})$. During training, the decoder recovers the watermark from these generated and warped frames. For I2V generation, we adopt AnimateLCM~\cite{animatelcm} as the video diffusion proxy. Unlike models such as SVD~\cite{stablevideodiffusion}, AnimateLCM attains high-quality videos in as few as two to four steps without classifier-free guidance (CFG)~\cite{cfg}, eliminating extra unconditional and conditional passes and reducing memory usage. Because it avoids CFG and requires only a handful of diffusion steps, it keeps the additional compute minimal while preserving temporal coherence. We generate \(M\) key frames from the proxy, including the \(0\)-th frame, and apply the above flow-guided warping before decoding.

\subsection{Watermarking Decoder}
The watermark decoder $D$ is a stride downsampling residual CNN that maps an image to a $k$ dimensional logit vector. It first encodes with strided convolutions until the spatial size matches the bit grid side $m=\sqrt{k}$, applies a stack of dilated residual blocks to aggregate mid and large contexts, and outputs a $1\times m\times m$ map whose flattened responses form the logits. We recover the watermark $k$ bits via:
\begin{equation}
w_q = \mathbf{1}\{\sigma(o_{q})>\tfrac12\} \in \{0,1\}^k, \quad
o_q = D(q) \in \mathbb{R}^k,
\label{eq:wmextract}
\end{equation}
\noindent where $q$ denotes any input to the decoder $D$, $\sigma(\cdot)$ is the sigmoid function, and $w_q$ is the recovered watermark.

\subsection{Optimization Strategy}
\label{optimization}
In our methodology we employ two main objective functions, one that optimizes the watermark encoder and decoder and another that optimizes the adversarial network.
\begin{figure*}
\centering
\includegraphics[width=\linewidth]{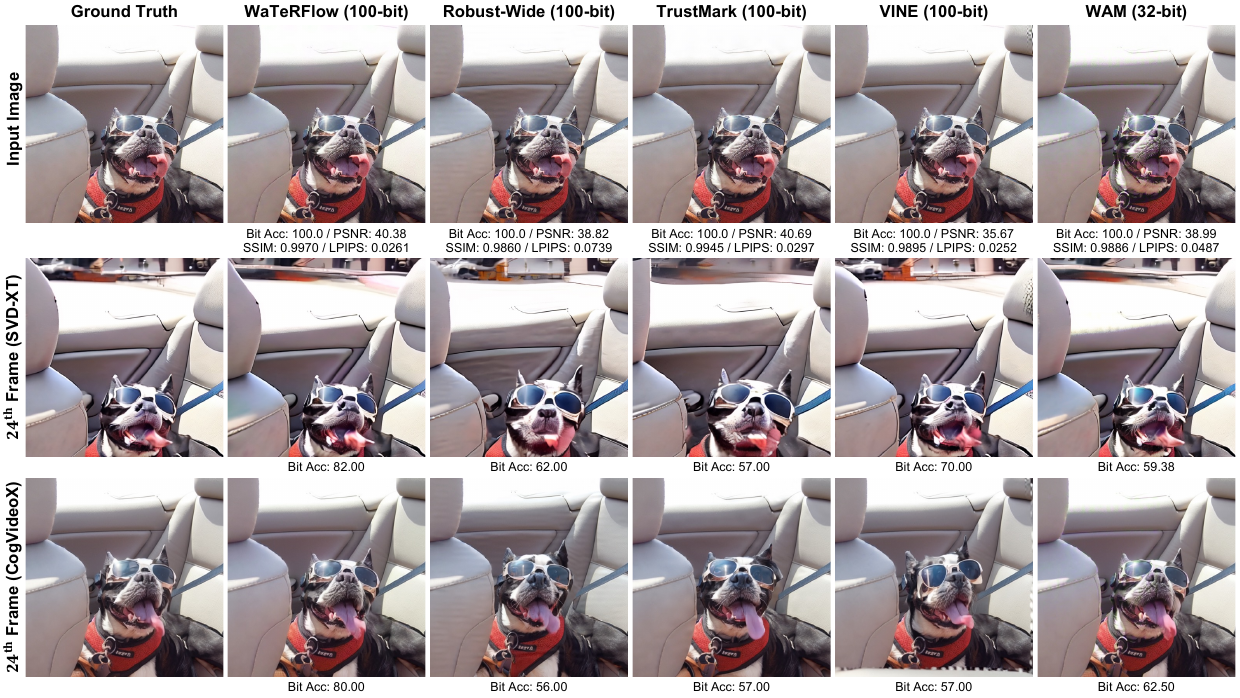}
\vspace{-19pt}
\caption{\textbf{Qualitative results.} Top: The original image and the watermarked images for each watermarking method. Middle: From left to right, the 24-th frames generated using SVD-XT are shown for the original image, our method, and the baselines. Bottom: Frames generated by CogVideoX. From left to right, we present the 24-th frames from the videos generated from the original image, then our method, followed by the baselines. Our method shows the highest bit accuracy for both video generation models in the given frames.}
\label{fig3}
\vspace{-15pt}
\end{figure*}
\myparagraph{Temporal Consistency Loss.}
I2V generation induces temporally varying subpixel drifts and misalignments that can persist even after flow-guided warping. These effects cause per-frame watermark predictions to fluctuate and degrade both average and late-frame bit accuracy. To stabilize decoding over time, we introduce a Temporal Consistency Loss (TCL) that encourages the decoder outputs for adjacent warped frames to be similar. Let $a_\ell$ be the $\ell$-th frame warped toward the first frame $v_0$ using the optical-flow estimator described in Eq.~\ref{eq:flow}. The TCL is defined as:
\begin{equation}
\mathcal{L}_{\text{TCL}}=\frac{1}{M-1}\sum_{\ell=1}^{M-1}\bigl\|D(a_{\ell})-D(a_{\ell-1})\bigr\|_2^2,
\label{eq:tcl}
\end{equation}
where $M$ is the number of key frames produced by FUSE, and $D$ is the watermark decoder.

\myparagraph{Encoder Loss.}
The watermark must be imperceptibly and stealthily embedded in the pixel space. Moreover, image editing and video generation operations are carried out in the latent space mapped by the VAE~\cite{vae} encoder, so the watermark must remain stealthy there as well. Accordingly, we define the encoder loss as follows:
\begin{equation}
\begin{split}
\mathcal{L}_{\text{enc}}
= \mathcal{L}_{\text{pixel}}
&+ \lambda_{\text{latent}}\,\mathcal{L}_{\text{latent}} \\
&+ \lambda_{\text{LPIPS}}\,\mathcal{L}_{\text{LPIPS}}
+ \lambda_{\text{sem}}\,\mathcal{L}_{\text{sem}},
\label{eq:encloss}
\end{split}
\end{equation}
\noindent where $\mathcal{L}_{\text{pixel}}$ denotes the mean squared error (MSE) between the original and watermarked image. $\mathcal{L}_{\text{latent}}$ denotes the MSE between $I$ and $I_w$ computed in the latent space mapped by VAE encoder.
$\mathcal{L}_{\text{LPIPS}}$ is the perceptual loss based on the LPIPS~\cite{lpips} evaluated between $I$ and $I_w$. Finally, $\mathcal{L}_{\text{sem}}$ is the semantic preservation loss defined in Eq.~\ref{eq:cliploss}.
\begin{figure*}[t]
\centering
\begin{subfigure}[t]{0.33\linewidth}
  \includegraphics[width=\linewidth]{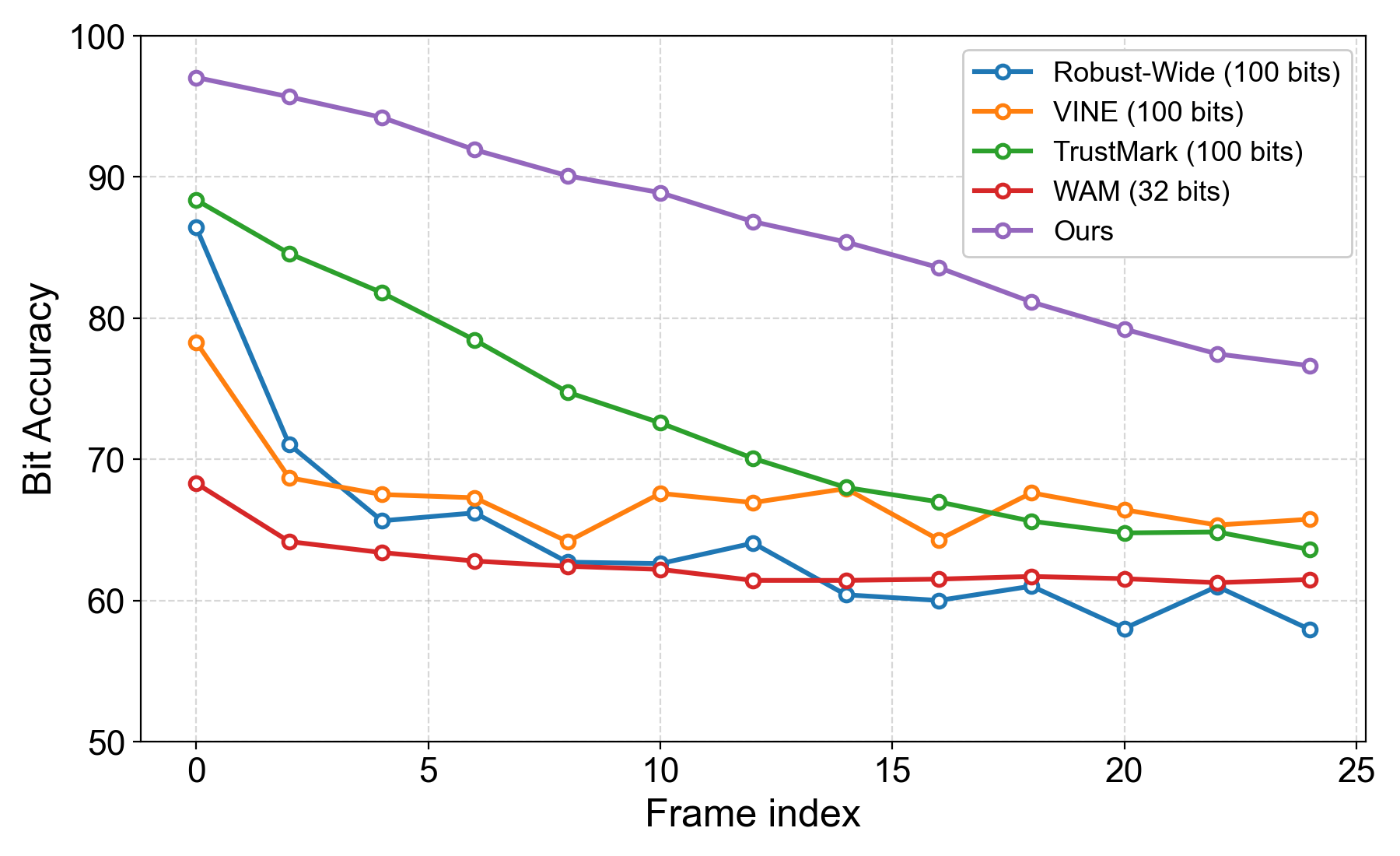}
  \caption{I2V with SVD-XT}
  \label{fig:svd_xt}
\end{subfigure}
\hfill
\begin{subfigure}[t]{0.33\linewidth}
  \includegraphics[width=\linewidth]{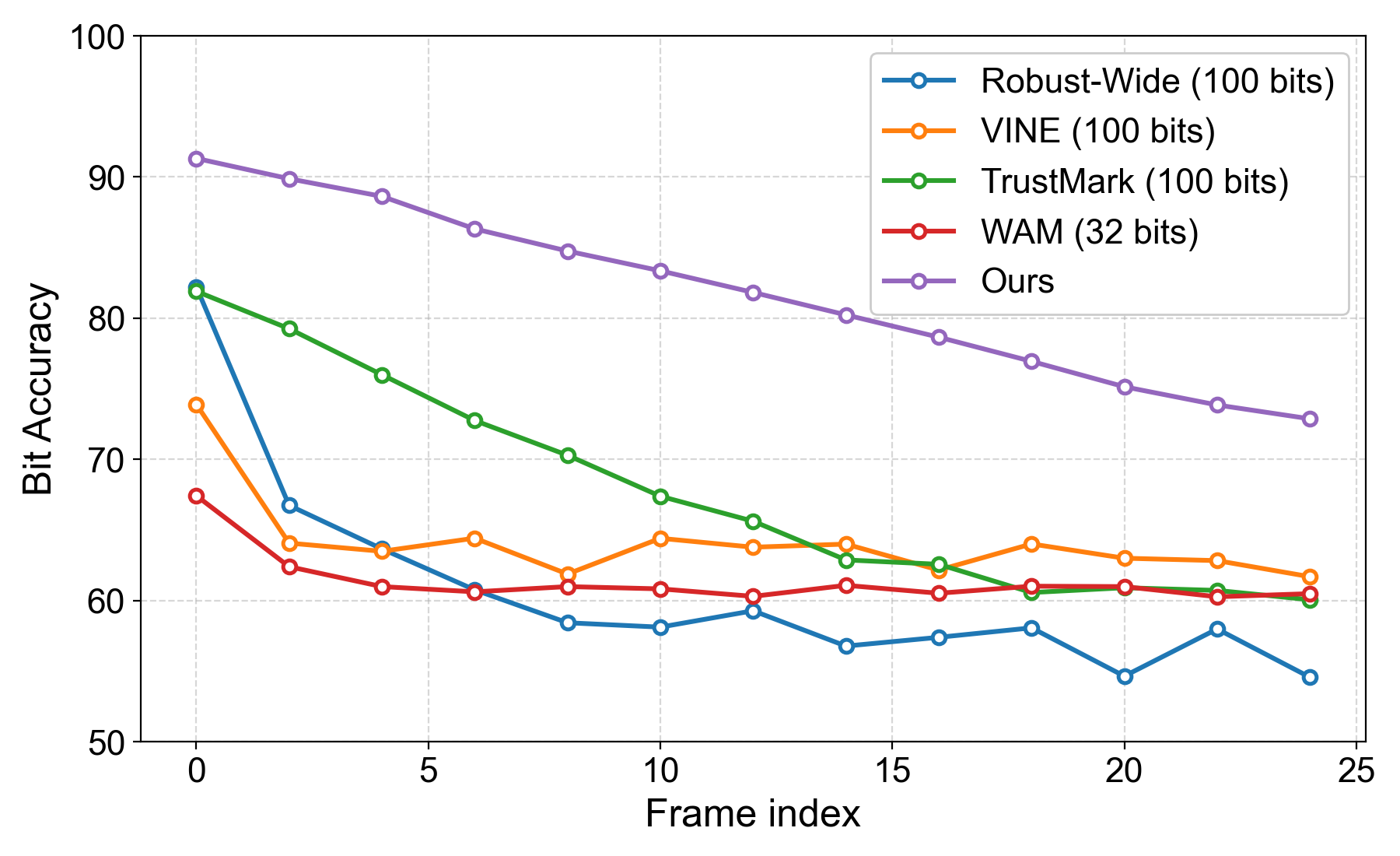}
  \caption{I2V with SVD-XT after image editing}
  \label{fig:i2v_edit}
\end{subfigure}
\hfill
\begin{subfigure}[t]{0.33\linewidth}
  \includegraphics[width=\linewidth]{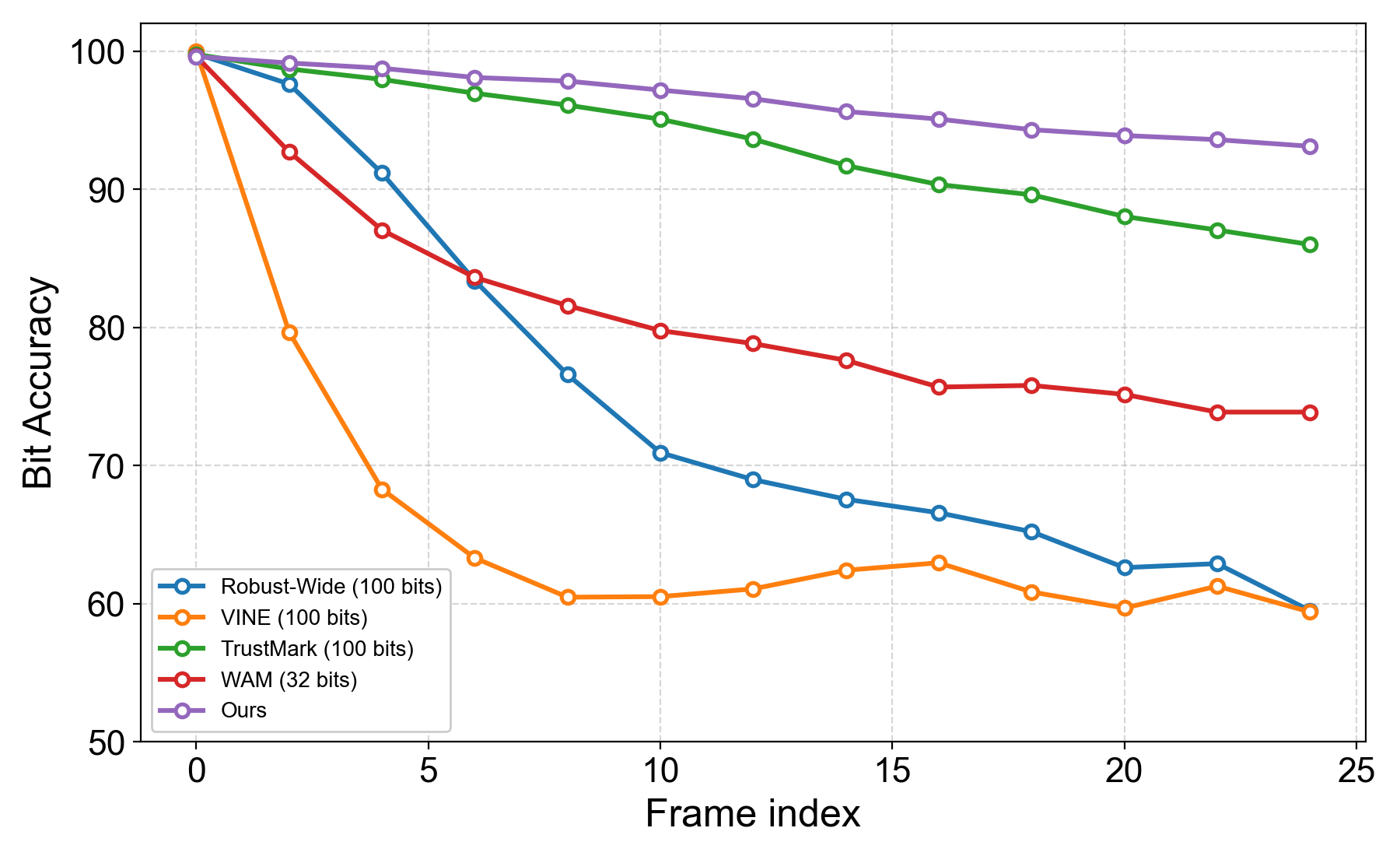}
  \caption{I2V with CogVideoX}
  \label{fig:cogvideox}
\end{subfigure}
\vspace{-8pt}
\caption{\textbf{Per-frame bit accuracy and I2V robustness.} Each plot visualizes bit accuracy on the even-numbered frames after image-to-video (I2V) generation. Across two representative I2V models, our method achieves the highest average bit accuracy compared to the baselines. It also shows the strongest robustness in the image-to-video (I2V) generation following image editing~\cite{ultra_edit}.}
\label{fig4}
\vspace{-5pt}
\end{figure*}
\newlength{\methodw}
\settowidth{\methodw}{Robust\text{-}Wide [15]}
\newlength{\capw}\settowidth{\capw}{100}
\begin{table*}[!t]
\centering
\small
\resizebox{\textwidth}{!}{%
\begin{tabular}{l c *{10}{c}}
\toprule
\multicolumn{2}{c}{} & \multicolumn{10}{c}{Bit Acc.(\%) $\uparrow$} \\
\cmidrule(lr){3-12}
% ↓ 1–2열을 헤더 2행에 걸쳐 세로 중앙 배치
\multirow{2}{*}{\makebox[\methodw][c]{Method}} &
\multirow{2}{*}{\makebox[\capw][c]{\makecell{Cap.\\[-0.25ex]\scriptsize (Bits)}}} &
\multicolumn{1}{c}{} & \multicolumn{9}{c}{Image-to-Video} \\
\cmidrule(lr){4-12}
& &
\shortstack{None\\\vphantom{$t_s=150$}} &
\shortstack{None\\\vphantom{$t_s=150$}} &
\shortstack{Editing\\\vphantom{$t_s=150$}} &
\shortstack{H.264\\$\text{CRF}=23$} &
\shortstack{Regeneration\\ $t_s=150$} &
\shortstack{Contrast\\ $\alpha=1.2$} &
\shortstack{Brightness\\ $s=1.2$} &
\shortstack{JPEG\\ $Q=50$} &
\shortstack{Gaussian Noise\\ $\sigma=0.05$} &
\shortstack{Gaussian Blur\\ $\sigma=1.5$} \\
\midrule
Robust-Wide~\cite{robustwide}  & 100 & \textbf{99.99} & 63.40 & 59.92 & 63.28 & 60.25 & 62.52 & 62.45 & 65.02 & 64.17 & 60.31 \\
VINE~\cite{vine}               & 100 & \textbf{99.99} & \underline{73.92} & 66.56 & \underline{73.75} & \underline{71.83} & \underline{79.72} & \underline{78.08} & \underline{75.15} & \underline{79.40} & \underline{77.66} \\
TrustMark~\cite{trustmark}     & 100 & \underline{99.92} & 73.76 & \underline{69.05} & 73.31 & 53.48 & 71.64 & 71.66 & 71.95 & 69.97 & 76.80 \\
WAM~\cite{wam}                 & 32  & 99.90 & 63.87 & 61.55 & 63.27 & 60.72 & 62.64 & 62.80 & 62.34 & 61.12 & 63.00 \\
\rowcolor{rowgray} WaTeRFlow (Ours)        & 100 & 99.86 & \textbf{84.96} & \textbf{80.20} & \textbf{84.64} & \textbf{81.98} & \textbf{81.68} & \textbf{81.45} & \textbf{81.14} & \textbf{82.73} & \textbf{85.78} \\
\bottomrule
\end{tabular}
}
\vspace{-5pt}
\caption{\textbf{Quantitative results.} The first None denotes the bit accuracy measured on the watermarked image with no preprocessing applied. All other columns report the \textit{average bit accuracy over frames} generated by SVD-XT. The H.264 column reports average bit accuracy measured after H.264 compression to the generated videos. The entries from Editing~\cite{ultra_edit} through Gaussian Blur indicate the preprocessing applied to the watermarked image before video generation. The best and second-best results are shown in \textbf{bold} and \underline{underlined}, respectively.
}
\label{tab1}
\vspace{-10pt}
\end{table*}

\myparagraph{Decoder Loss.}
The decoder is trained for per-bit message recovery on the watermarked image, the edited watermarked image, and flow-warped video frames. We adopt binary cross entropy (BCE) on per-bit probabilities:
\begin{equation}
\begin{split}
\text{BCE}(x,y)=-\frac{1}{L}\sum_{i=1}^L&\big[y_i\log x_i\\
&+(1-y_i)\log(1-x_i)\big],
\label{eq:bce}
\end{split}
\end{equation}
where $x\in\mathbb{R}^{L}$ is the predicted probability vector and $y\in\mathbb{R}^{L}$ is the binary target. The decoder loss is defined as:
\begin{subequations}\label{eq:decloss}
\begin{alignat}{2}
& \mathcal{L}_{\text{dec}} &&= \mathcal{L}_{\text{TCL}}+\mathcal{L}_{\text{MSG}}, \label{eq:dec_loss_a} \\
& \mathcal{L}_{\text{MSG}} &&= \frac{1}{M}\sum_{\ell=0}^{M-1}\mathrm{BCE}\!\left(p_{a_\ell}, w\right) \notag\\
& &&\quad + \mathrm{BCE}(p_{I_w}, w) + \mathrm{BCE}(p_{\tilde{I}_w}, w), \label{eq:msg_loss_a}
\end{alignat}
\end{subequations}
where $\mathcal{L}_{\text{TCL}}$ is defined in Eq.~\ref{eq:tcl} and $p_{I_w}$, $p_{\tilde I_w}$, and $p_{a_\ell}$ are the per-bit probabilities obtained by applying the sigmoid function to the decoder outputs for $I_w$, the edited image $\tilde I_w$ from FUSE’s editing branch, and the $\ell$-th flow-warped frame $a_\ell$.

\myparagraph{Adversarial Training.}
To enable the encoder to embed the watermark signal while preserving image quality, we adopt an adversarial strategy in which an adversarial network $A$ is trained to distinguish original images from watermarked ones, and the encoder $E$ learns to fool $A$, thereby encouraging quality aware watermark embeddings. Following prior work~\cite{usingpatchgan}, we use a PatchGAN~\cite{patchgan} adversarial network that outputs a patch-wise logit map, providing dense supervision and promoting imperceptible embeddings. The adversarial loss is defined as follows:
\begin{equation}
\mathcal{L}_{\text{adv}}^{G}=-\frac{1}{H'W'}\sum_{i=1}^{H'}\sum_{j=1}^{W'}\log \sigma\!\big(A(I_w)_{i,j}\big),
\label{eq:advG}
\end{equation}
\noindent where \(A(I_w)_{i,j}\) denotes the \((i,j)\)-th logit of the adversarial network $A$ for the watermarked image, and $H'$, $W'$ are the adversarial network output map sizes. Meanwhile, the loss for the adversarial network \(A\) itself is defined as follows:
\begin{equation}
\begin{split}
\mathcal{L}_{\text{disc}}=
-\frac{1}{H'W'}
\sum_{i,j}
&\Big[
\log \sigma\big(A(I)_{i,j}\big) \\
&+\log \big(1 - \sigma\big(A(I_{w})_{i,j}\big)\big)
\Big].
\end{split}
\label{eq:discloss}
\end{equation}
\noindent This objective encourages the adversarial network to assign high probabilities to patches from the original image and low probabilities to patches from the watermarked image.

\myparagraph{Total Loss.}
The overall objective function for training our proposed watermark encoder and decoder is given by:
\begin{equation}
\mathcal{L}_{\text{total}}=
\mathcal{L}_{\text{enc}}
+ \lambda_{\text{dec}}\mathcal{L}_{\text{dec}}
+ \lambda_{\text{adv}}\mathcal{L}_{\text{adv}}^{G}.
\label{eq:totalloss}
\end{equation}
\section{Experiments}
\label{sec:exp}
This section presents experimental analysis, including details of the experimental setup, evaluation, comparisons with baseline methods, and ablation experiments.

\subsection{Experimental Setting}
\myparagraph{Dataset.}
We train our models using the dataset proposed in InstructPix2Pix~\cite{instructpix2pix}. From this dataset, we use 20{,}000 original image--prompt pairs for training, and we use the paired textual instructions as editing prompts in the image editing branch of our framework.  For evaluation, we use real-world images at a resolution of \(512\times512\) drawn from UltraEdit~\cite{ultra_edit}, from which we randomly sample 500 to assess robustness in I2V generation for watermarked images. The graphs presented in Fig.~\ref{fig4} are produced using a randomly selected subset of 100 images from those 500. Meanwhile, the text prompts required for the CogVideoX experiments are generated with Qwen~2.5-VL-7B-Instruct~\cite{qwen}. Please refer to the Supplementary Material for further details.

\myparagraph{Baseline.}
We use the officially released checkpoints and training code to standardize each model's capacity to 100-bit as much as possible. The only exception is WAM~\cite{wam}. Its methodology explicitly states a 32-bit capacity limit, so we retain its original capacity. Additionally, because the baseline models were trained at different image resolutions, we follow the resolution-scaling procedure introduced in VINE~\cite{vine} to evaluate all models at a common resolution.

\myparagraph{Evaluation Metrics.}
We evaluate image watermarking methods along two dimensions. \textit{Robustness}: We measure robustness using bit accuracy, defined as the fraction of the embedded watermark that is correctly recovered by the watermark decoder from a watermarked image or from frames of a generated video. Beyond using watermarked images in the image-to-video (I2V) generation, we conduct robustness experiments that apply distortions immediately before or after video generation. The distortions include image editing~\cite{ultra_edit}, H.264 ($\text{CRF}=23$), regeneration with a VP~\cite{diffusion_model} scheduler ($t_s=150$), contrast adjustment ($\alpha=1.2$), brightness adjustment ($s=1.2$), JPEG compression ($Q=50$), Gaussian noise ($\sigma=0.05$), and Gaussian blur ($\sigma=1.5$). For example, when applying the regeneration, we first run the regeneration procedure on the watermarked image, then perform I2V generation, and measure bit accuracy. \textit{Imperceptibility}: We assess image quality between the original and watermarked images using PSNR, SSIM~\cite{ssim}, and LPIPS~\cite{lpips}. Using the same metrics, we compare frames generated from original images with those from watermarked images to assess video quality.

\myparagraph{Implementation Details.}
We train on a single A100 GPU for 20{,}000 steps. In FUSE, all pretrained components remain frozen during training, as described in Sec.~\ref{subsec:fuse}. FUSE generates four frames, including frame 0, with \(M\) denoting the count in Eq.~\ref{eq:tcl}. Starting at step 7{,}000, we activate FUSE’s video generation branch, and in Eq.~\ref{eq:decloss} the BCE terms and TCL are computed on flow-warped frames. We optimize Eq.~\ref{eq:totalloss} with \(\lambda_{\text{latent}}=10^{-3}\), \(\lambda_{\text{LPIPS}}=0.18\), \(\lambda_{\text{sem}}=10^{-3}\), \(\lambda_{\text{dec}}=1.3\), and \(\lambda_{\text{adv}}=0.004\). During training, FUSE uses InstructPix2Pix~\cite{instructpix2pix} and AnimateLCM~\cite{animatelcm} as proxy models. At evaluation, we generate frames with SVD-XT~\cite{stablevideodiffusion}, a U-Net based model, and CogVideoX~\cite{cogvideox}, a DiT based model, to demonstrate robustness across architectures. We then perform image edits with UltraEdit~\cite{ultra_edit}, treating it as one of the pre-I2V distortion scenarios. Specifically, we first apply UltraEdit to the watermarked image and then generate a video with SVD-XT~\cite{stablevideodiffusion}.

\subsection{Experimental Results}
\myparagraph{Image-to-Video Robustness.}
In this section, we evaluate the robustness of the watermark under I2V generation and compare our method against baselines. Using watermarked images as inputs, we generate videos with two models, SVD-XT~\cite{stablevideodiffusion} and CogVideoX~\cite{cogvideox}. As shown in Fig.~\ref{fig:svd_xt}, our method achieves the highest bit accuracy across all evaluated frames, including the first generated frame. Likewise, as shown in Fig.~\ref{fig:cogvideox}, our method attains the highest average bit accuracy over the evaluated frames for CogVideoX compared to the baselines. Overall, across both a U-Net based video generator~\cite{stablevideodiffusion} and a DiT based video generator~\cite{cogvideox}, our method exhibits greater robustness than the baselines. We conducted additional experiments and analyses on the relationship between the extent of motion in the generated videos and the bit accuracy in later frames. Please refer to the Supplementary Material for details and results.

\myparagraph{More Challenging Scenarios.}
We evaluate robustness against distortions applied to the watermarked image before I2V and to the generated video after I2V. We consider a more demanding setting where the watermarked image is first modified by an image-editing model~\cite{ultra_edit} and the edited image is then used as the input to the video generator. In addition, we include common degradations such as JPEG compression prior to video generation. We also evaluate robustness when videos generated from watermarked images are encoded with H.264 compression. As seen in Fig.~\ref{fig:i2v_edit} and Tab.~\ref{tab1}, our method is more robust than the baselines for I2V under pre-I2V distortions including image editing, regeneration, contrast and brightness changes, JPEG compression, Gaussian noise, and Gaussian blur. It also remains robust under H.264 compression applied after I2V.

\myparagraph{Imperceptibility Evaluation.}
We evaluate the quality of watermarked images and the frames generated from watermarked images. Even when considering the trade-offs among robustness, imperceptibility, and capacity, Fig.~\ref{fig3} and Tab.~\ref{tab:image-video-quality} show that our method yields competitive quality. In particular, as shown in Fig.~\ref{fig3}, our method maintains motion close to the original while achieving higher bit accuracy.
% \begin{table}[t]
% \centering
% \small
% \renewcommand{\arraystretch}{1.12}
% \setlength{\extrarowheight}{1pt}
% \resizebox{\columnwidth}{!}{
% \begin{tabular}{l c c c c c c c c}
% \toprule
% \multicolumn{1}{c}{\multirow[c]{2}{*}{Method}} &
% \multirow[c]{2}{*}{\makecell{Cap.\\(Bits)}} &
% \multicolumn{3}{c}{Watermarked Image} & \multicolumn{4}{c}{Generated Video} \\
% \cmidrule(lr){3-5}\cmidrule(lr){6-9}
% & & PSNR $\uparrow$ & SSIM $\uparrow$ & LPIPS $\downarrow$
%   & PSNR $\uparrow$ & SSIM $\uparrow$ & LPIPS $\downarrow$ & Bit Acc. (\%) $\uparrow$ \\
% \midrule
% Robust-Wide~\cite{robustwide} & 100 & 36.88 & 0.9698 & 0.0649 & 19.56 & 0.7228 & 0.2884 & 63.40 \\
% VINE~\cite{vine}              & 100 & 33.88 &\underline{0.9828} & \underline{0.0334} & 17.86 & 0.6879 & 0.2930 & \underline{73.92} \\
% TrustMark~\cite{trustmark}    & 100 & \textbf{40.33} & \textbf{0.9902} & 0.0348 & \underline{22.37} & \textbf{0.7982} & \underline{0.1867} & 73.76 \\
% WAM~\cite{wam}                & 32  & 37.91 & 0.9796 & 0.0337 & 20.51 & 0.7527 & 0.2232 & 63.87 \\
% \rowcolor{rowgray} WaTeRFlow (Ours)       & 100 & \underline{38.33} & \textbf{0.9902} & \textbf{0.0291} & \textbf{22.43} & \underline{0.7910} & \textbf{0.1816} & \textbf{84.96} \\
% \bottomrule
% \end{tabular}
% }
% \vspace{-5pt}
% \caption{\textbf{Quality of watermarked images and generated videos.} Except for WAM, the watermark capacity is fixed at 100 bits. The quality metrics for generated videos are evaluated against videos generated from images without embedded watermarks.}
% \label{tab:image-video-quality}
% \vspace{-5pt}
% \end{table}
\begin{table}[t]
\centering
\small
\renewcommand{\arraystretch}{1.12}
\setlength{\extrarowheight}{1pt}
\resizebox{\columnwidth}{!}{
\begin{tabular}{l c c c c c c c c c}
\toprule
\multicolumn{1}{c}{\multirow[c]{2}{*}{Method}} &
\multirow[c]{2}{*}{\makecell{Cap.\\(Bits)}} &
\multicolumn{4}{c}{Watermarked Image} &
\multicolumn{4}{c}{Generated Video} \\
\cmidrule(lr){3-6}\cmidrule(lr){7-10}
& & PSNR $\uparrow$ & SSIM $\uparrow$ & LPIPS $\downarrow$ & Bit Acc. (\%)$\uparrow$
  & PSNR $\uparrow$ & SSIM $\uparrow$ & LPIPS $\downarrow$ & Bit Acc. (\%)$\uparrow$ \\
\midrule
Robust-Wide~\cite{robustwide} & 100 & 36.88 & 0.9698 & 0.0649 & \textbf{99.99} & 19.56 & 0.7228 & 0.2884 & 63.40 \\
VINE~\cite{vine}              & 100 & 33.88 & \underline{0.9828} & \underline{0.0334} & \textbf{99.99} & 17.86 & 0.6879 & 0.2930 & \underline{73.92} \\
TrustMark~\cite{trustmark}    & 100 & \textbf{40.33} & \textbf{0.9902} & 0.0348 & \underline{99.92} & \underline{22.37} & \textbf{0.7982} & \underline{0.1867} & 73.76 \\
WAM~\cite{wam}                & 32  & 37.91 & 0.9796 & 0.0337 & 99.90 & 20.51 & 0.7527 & 0.2232 & 63.87 \\
\rowcolor{rowgray} WaTeRFlow (Ours) & 100 & \underline{38.33} & \textbf{0.9902} & \textbf{0.0291} & 99.86 & \textbf{22.43} & \underline{0.7910} & \textbf{0.1816} & \textbf{84.96} \\
\bottomrule
\end{tabular}
}
\vspace{-5pt}
\caption{\textbf{Quality of watermarked images and generated videos.} The quality metrics for generated videos are evaluated against videos generated from images without embedded watermarks. In this setting, the videos are generated using SVD-XT.}
\label{tab:image-video-quality}
\vspace{-5pt}
\end{table}

\begin{table}[!t]
\centering
\resizebox{\columnwidth}{!}{%
\begin{tabular}{cccc|ccc|ccc}
\toprule
\multicolumn{2}{c}{FUSE} & \multirow[c]{2}{*}{$\mathcal{L}_{\text{TCL}}$} & \multirow[c]{2}{*}{$\mathcal{L}_{\text{sem}}$} & \multicolumn{3}{c}{Bit Acc.(\%) $\uparrow$} & \multicolumn{3}{c}{Watermarked image} \\
\cmidrule(lr){1-2}\cmidrule(lr){5-7}\cmidrule(lr){8-10}
I & V & & & First frame & Avg. & Avg. w/ Edit & PSNR $\uparrow$ & SSIM $\uparrow$ & LPIPS $\downarrow$ \\
\midrule
- & - & - & -                                  & 55.12 & 55.11 & 54.87 & \textbf{52.65} & \textbf{0.9972} & \textbf{0.0002} \\
-          & \checkmark & \checkmark & \checkmark   & \underline{93.26} & \underline{81.65} & \underline{77.47} & 36.28 & 0.9868 & 0.0422 \\
\checkmark & -          & -          & \checkmark   & 91.89 & 66.32 & 63.56 & 39.25 & 0.9898 & \underline{0.0262} \\
\checkmark & \checkmark & -          & \checkmark   & 92.23 & 79.11 & 74.67 & 36.91 & 0.9860 & 0.0462 \\
\checkmark & \checkmark & \checkmark & -            & 89.63 & 76.82 & 72.79 & \underline{40.44} & \underline{0.9920} & 0.0299 \\
\rowcolor{rowgray}\checkmark & \checkmark & \checkmark & \checkmark
& \textbf{96.93} & \textbf{84.96} & \textbf{80.20} & 38.33 & 0.9902 & 0.0291 \\
\bottomrule
\end{tabular}
}% end resizebox
\vspace{-5pt}
\caption{\textbf{Ablation of WaTeRFlow.} I and V denote the image-editing and video generation branches, and $\mathcal{L}_{\text{TCL}}$ and $\mathcal{L}_{\text{sem}}$ denote the loss terms in Sec.~\ref{sec:method}. Avg. indicates the average bit accuracy over frames generated by SVD-XT, while Avg. w/ Edit denotes the average bit accuracy after editing the watermarked image.}
\label{tab4}
\vspace{-15pt}
\end{table}

\subsection{Ablation Study}
\label{sec:ablation}
\myparagraph{FUSE, TCL, and Semantic Preservation Loss.}
In this section, we evaluate the impact of FUSE, TCL, and the Semantic Preservation loss on robustness under I2V generation. With all other settings fixed, we ablate one component at a time during training. Using the resulting encoders, we then generate SVD-XT videos from watermarked images and measure per-frame bit accuracy. The results are summarized in Tab.~\ref{tab4}, and the average accuracy is highest when all three components are enabled. FUSE exposes the encoder and decoder to realistic distortions during training. TCL stabilizes frame-wise outputs after warping. The semantic preservation loss maintains the semantic information of the watermarked image and improves the accuracy of frame 0. Furthermore, Tab.~\ref{tab4} shows that removing FUSE’s image editing branch leads to a decrease in bit accuracy for frames generated after editing the watermarked image. For further details, please refer to the Supplementary Material.

\myparagraph{Video Diffusion Proxy Selection.}
We analyze how the choice of video diffusion proxy for FUSE’s video generation branch affects performance and training cost. Tab.~\ref{tab3} summarizes results when we use SVD, SVD without CFG, and AnimateLCM in the video generation branch. AnimateLCM yields faster video generation and requires less peak VRAM during training than SVD without CFG. Considering the overall performance of SVD-based training, using SVD effectively requires enabling CFG. Taking this into account, AnimateLCM achieves competitive performance with lower peak VRAM and shorter training time.
\begin{table}[ht]
\centering
\small
\setlength{\tabcolsep}{5pt}
\resizebox{\columnwidth}{!}{%
\begin{tabular}{l c c c c c c c}
\toprule
\multicolumn{1}{c}{Method} & Bit Acc. (\%) $\uparrow$ & PSNR $\uparrow$ & SSIM $\uparrow$ & LPIPS $\downarrow$ & VRAM (GB) & Train (h) & Gen. (s) \\
\midrule
SVD~\cite{stablevideodiffusion} (w/o CFG)        & 72.93 & 38.55 & 0.9883 & 0.0298 & 67.98 & 23.09 & 2.6 \\
SVD~\cite{stablevideodiffusion} (w/ CFG)         & \underline{83.83} & \textbf{38.99} & \textbf{0.9909} & \textbf{0.0254} & 93.80 & 42.39 & 7.9 \\
\rowcolor{rowgray} AnimateLCM~\cite{animatelcm} (Ours)
                     & \textbf{84.96} & \underline{38.83} & \underline{0.9902} & \underline{0.0291} & 44.51 & 17.70 & 1.1 \\
\bottomrule
\end{tabular}}
\vspace{-5pt}
\caption{\textbf{Ablation on video diffusion proxy choice.} We compare two SVD variants, one without CFG and one with CFG, together with AnimateLCM. Bit Acc. is measured on frames generated by SVD-XT, VRAM denotes peak memory usage during training, Train is total training time, and Gen. is per-video generation. Here, PSNR, SSIM, and LPIPS represent watermarked image quality.}
\label{tab3}
\vspace{-13pt}
\end{table}

\begin{figure}[ht]
\vspace{-5pt}
  \centering
  \includegraphics[width=\columnwidth]{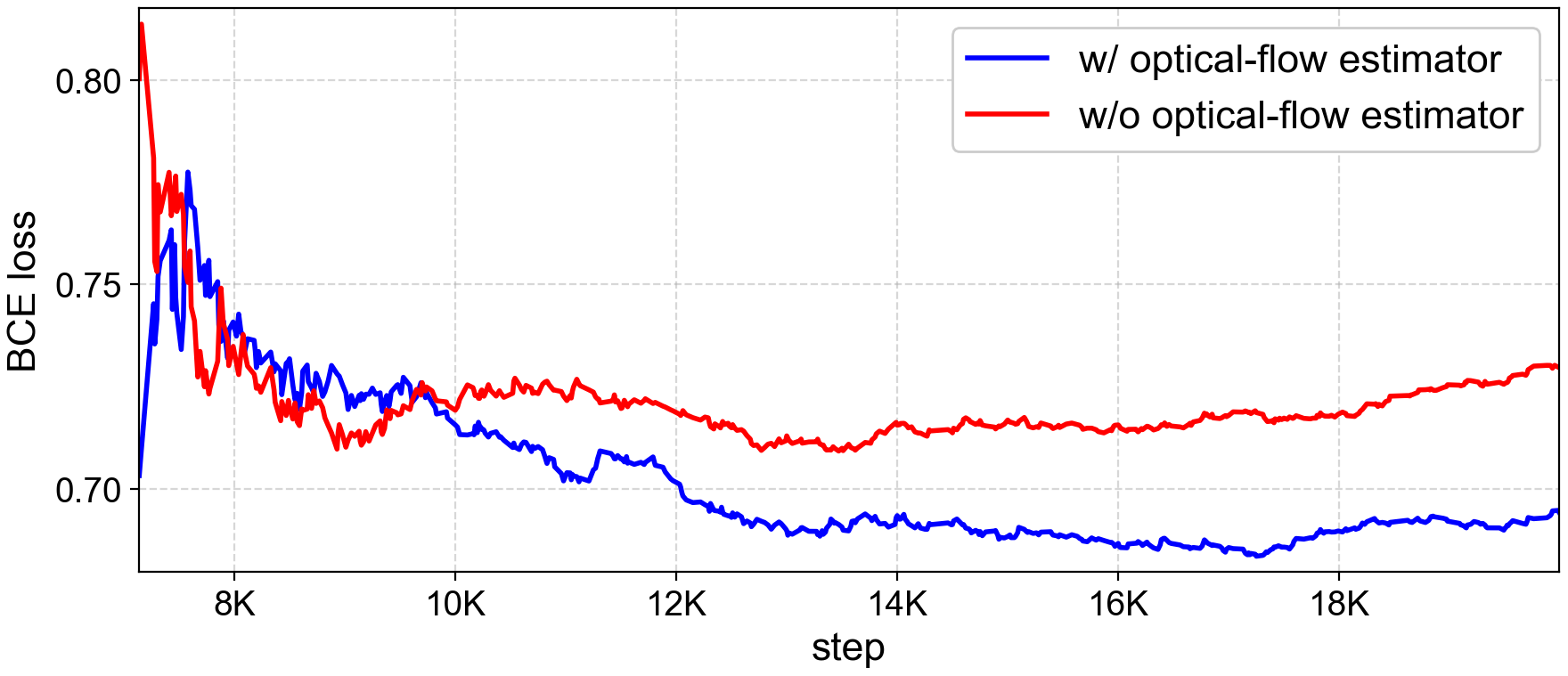}
  \vspace{-20pt}
  \caption{\textbf{Training stability.} Computed on frames from FUSE’s video generation branch, comparing with and without an optical-flow estimator. The estimator yields a more stable trajectory, indicating improved robustness to image-to-video generation.}
  \label{fig:fig5}
\vspace{-10pt}
\end{figure}

\myparagraph{Optical Flow and Training Stability.}
We analyze how the presence or absence of an optical-flow estimator in FUSE’s video generation branch affects training. Fig.~\ref{fig:fig5} visualizes the BCE loss measured on frames generated during training, which serves as an indicator of the decoder’s recovery performance. The graph suggests that without an optical- flow estimator, the decoder does not adequately adapt to distortions introduced by image-to-video generation.

\myparagraph{Semantic preservation with DINOv2.}
We replace the CLIP-based semantic preservation loss with a DINOv2~\cite{dinov2} based variant. It is defined in a representation space different from the CLIP conditioning encoder used by SVD. Despite this difference, constraining watermark insertion to preserve structural cues such as object layout, contours, and low frequency signals improves first-frame bit accuracy after image-to-video synthesis. The pixel-level and latent-level objectives in Eq.~\ref{eq:encloss} already shrink the overall residual, and the DINOv2-based term then guides where the remaining change resides by reducing structure sensitive changes and keeping the remaining change in local details with smaller impact. Considering the first-frame bit accuracy of the w/o $\mathcal{L}_{\text{sem}}$ setting in Tab.~\ref{tab4}, Tab.~\ref{tab5} shows that using DINOv2 also yields higher first-frame bit accuracy.
\begin{table}[ht]
\centering
\small
\resizebox{\columnwidth}{!}{
\begin{tabular}{l cc ccc}
\toprule
\multicolumn{1}{c}{\multirow{2}{*}{Method}} & \multicolumn{2}{c}{Bit Acc. (\%) $\uparrow$} & \multicolumn{3}{c}{Watermarked Image} \\
\cmidrule(lr){2-3}\cmidrule(lr){4-6}
& First frame & Avg. & PSNR $\uparrow$ & SSIM $\uparrow$ & LPIPS $\downarrow$ \\
\midrule
DINOv2~\cite{dinov2} & 95.84 & 83.49 & 38.27 & 0.9873 & 0.0406 \\
CLIP~\cite{clip} & 96.93 & 84.96 & 38.83 & 0.9902 & 0.0291 \\
\bottomrule
\end{tabular}
}
\vspace{-5pt}
\caption{\textbf{Ablation on semantic preservation loss.} Even when the CLIP is replaced with DINOv2, it improves bit accuracy on the first frame after image-to-video synthesis. Avg. denotes the average bit accuracy across frames generated by SVD-XT.}
\label{tab5}
\vspace{-15pt}
\end{table}
\section{Conclusion}
In this work, we study an image-to-video provenance scenario that current image watermarking methods rarely consider. A single watermarked image can be turned into a video and the watermark must be recovered from generated frames. We introduce WaTeRFlow, which trains encoder and decoder under realistic image-to-video generation while preserving watermarked image quality. WaTeRFlow relies on three key components. The Flow-guided Unified Synthesis Engine exposes the system to instruction-driven image editing and fast video diffusion during learning. A Temporal Consistency Loss stabilizes frame-level decoding and a semantic preservation loss keeps the conditioning signal close to that of the original image. WaTeRFlow improves bit accuracy for the first frame and subsequent frames while minimally affecting the visual quality of watermarked images, making it a practical tool for provenance and ownership verification.

\myparagraph{Limitations.}
We evaluate robustness on Stable Video Diffusion and CogVideoX. However, if the video generator substantially alters the initial frame or the semantic content of the conditioning signal, bit accuracy may drop. Robustness can also degrade for generators whose synthesis principles differ markedly from the video proxy models used during training. Although WaTeRFlow improves robustness in image-to-video scenarios, further discussion is needed regarding adversarial removal of the embedded watermark.

% \section{Conclusion}
% We address an I2V scenario overlooked by existing image watermarking—recovering a watermark when a single watermarked image is converted into a video by an I2V model. We introduce \textbf{WaTeRFlow}, which combines exposure during training to realistic edits and I2V distortions through the Flow-guided Unified Synthesis Engine (FUSE), a Temporal Consistency Loss (TCL) that stabilizes per-frame predictions, and a semantic preservation loss that maintains the conditioning signal. Across representative generators, conditioning on a watermarked image enables accurate recovery from synthesized videos. WaTeRFlow improves first frame and frame-wise bit accuracy, preserves imperceptibility, and mitigates the late-frame drop-off in watermark detection rates.

% \myparagraph{Limitations.}
% We verify robustness on Stable Video Diffusion (SVD), a video generator built on U-Net, and on the diffusion-transformer--based CogVideoX. However, if the video generator substantially alters the initial frame or its conditioning semantics, correspondences can weaken and detection may drop. Robustness may also degrade for generators whose synthesis principles differ markedly from the proxy models used during training. Finally, although WaTeRFlow enhances robustness in realistic I2V settings, further discussion is needed regarding deliberate adversarial removal.
\clearpage
{
    \small
    \bibliographystyle{ieeenat_fullname}
    \bibliography{main}
}

%\clearpage
%\appendix
%\input{_suppl}
%\clearpage
% {
%     \small
%     \bibliographystylesupp{ieeenat_fullname}
%     \bibliographysupp{main,_suppl}
% }
\end{document}

%%%%%%%%% TITLE
\renewcommand{\thesection}{\Alph{section}} 
\newcommand{\myparagraph}[1]{\vspace{4pt}\noindent{\bf #1}}

\title{WaTeRFlow: Watermark Temporal Robustness via Flow Consistency}

% \maketitle
\thispagestyle{empty}
\maketitlesupplementary

%%%%%%%%% SECTION INPUTS
\section*{Overview}
\label{overview}
This supplementary material presents additional experimental results and further details of our proposed method, WaTeRFlow.
\begin{itemize}
\item Sec.~\ref{additionalresults} presents additional experimental results of WaTeRFlow.
\item Sec.~\ref{additionalablationstudy} provides additional experiments and explanations for the ablation study.
\item Sec.~\ref{furtherexperimentaldetails} describes additional details of the experimental setup for the experiments presented in the main paper and the supplementary material.
\item Sec.~\ref{failurecase} provides the limitations of the proposed WaTeRFlow through failure cases.
\item Sec.~\ref{additionalqualitativeresults} provides additional qualitative evaluation results.
\end{itemize}
\section{Additional Results}
\label{additionalresults}
\subsection{Robustness to Video Dynamics}
\begin{figure*}[t]
\vspace{-5pt}
\centering
\begin{subfigure}[t]{0.32\textwidth}
  \centering
  \includegraphics[width=\linewidth]{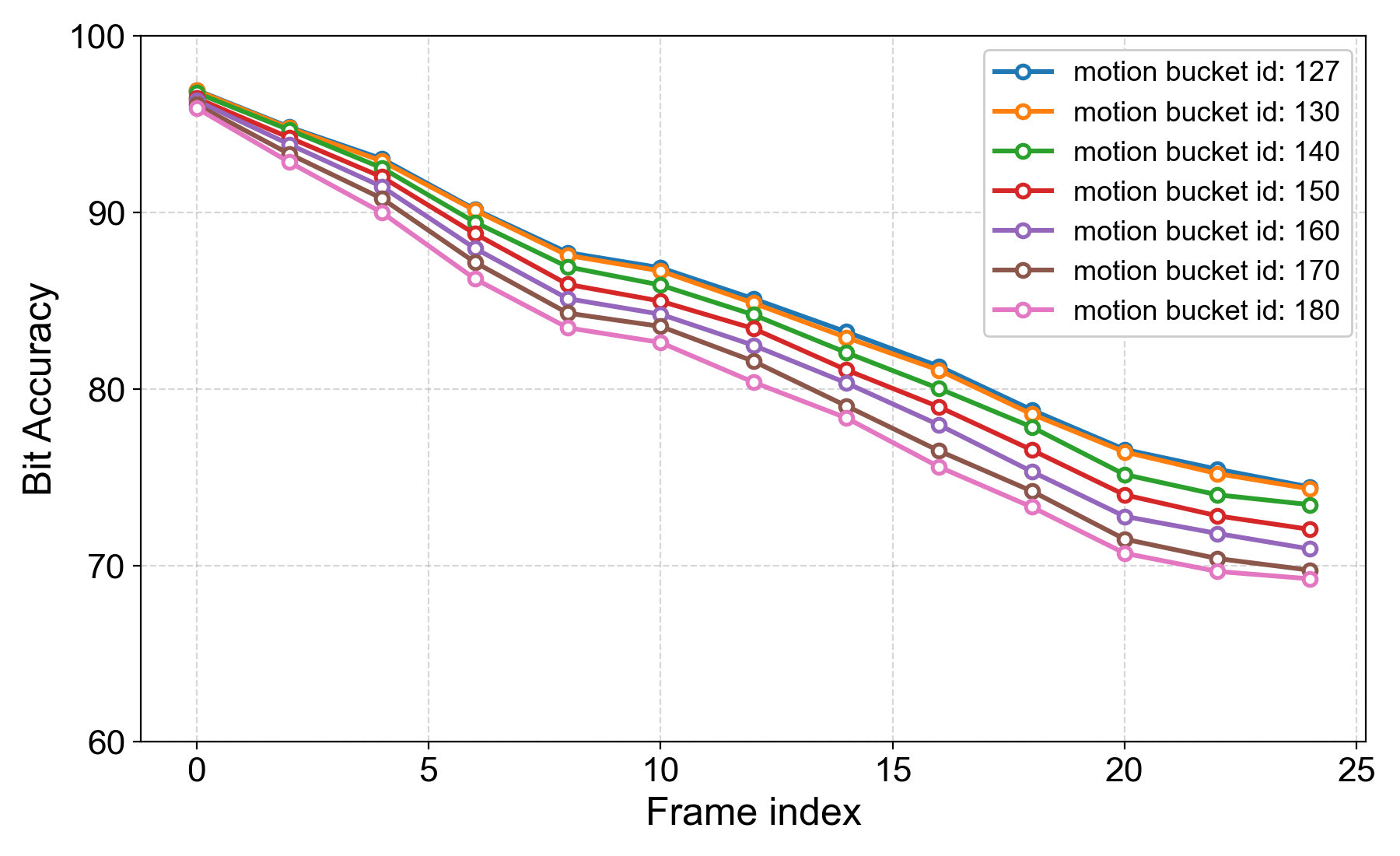}
  \caption{Bit accuracy of WaTeRFlow on SVD-XT vs. \texttt{motion\_bucket\_id}}
  \label{supfig:fig1}
\end{subfigure}
\hfill
\begin{subfigure}[t]{0.32\textwidth}
  \centering
  \includegraphics[width=\linewidth]{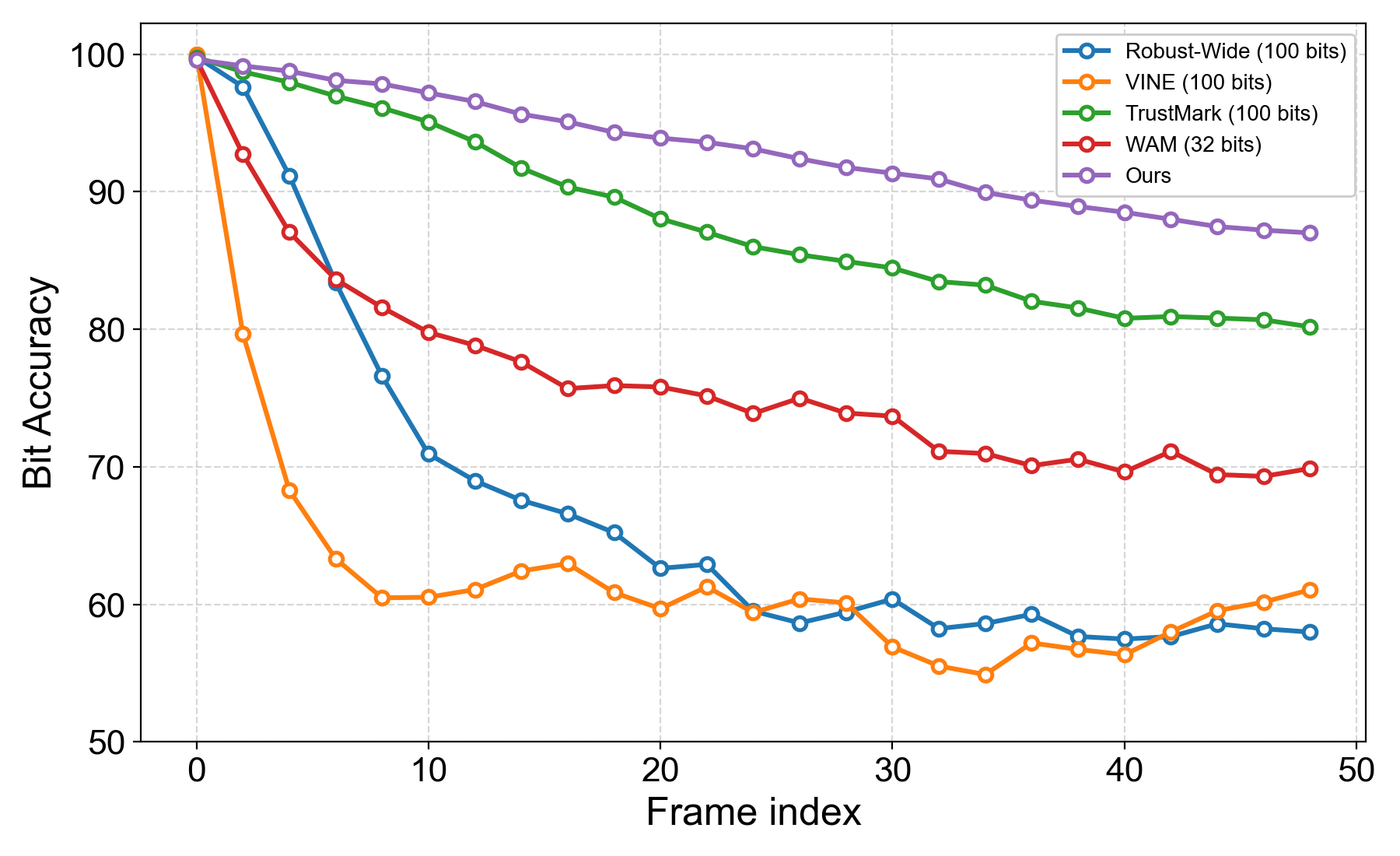}
  \caption{I2V with CogVideoX}
  \label{supfig:fig2}
\end{subfigure}
\hfill
\begin{subfigure}[t]{0.32\textwidth}
  \centering
  \includegraphics[width=\linewidth]{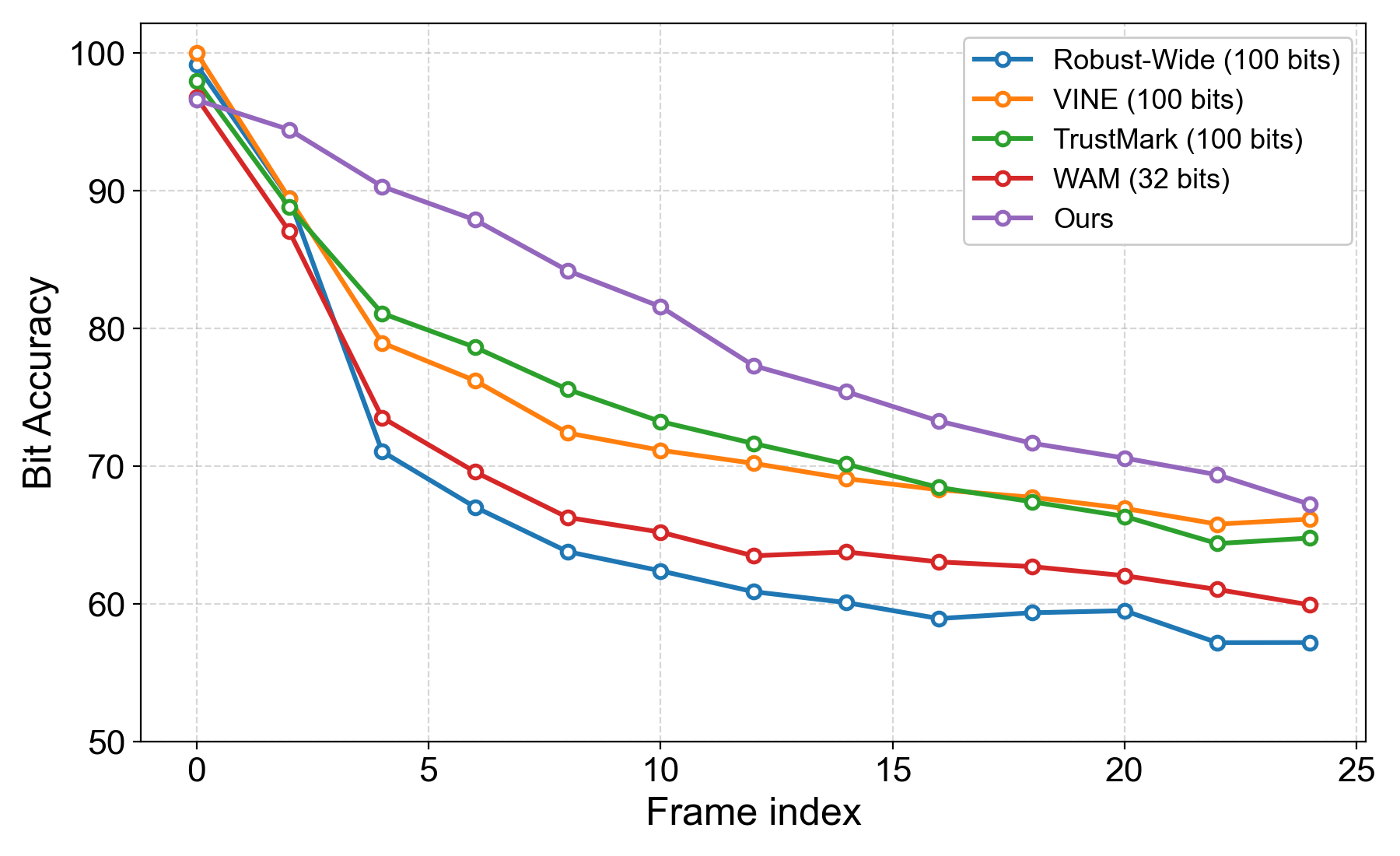}
  \caption{I2V with Vidu~\cite{vidu}}
  \label{supfig:fig3}
\end{subfigure}
\vspace{-5pt}
\caption{\textbf{Per-frame bit accuracy and I2V robustness.} Each plot shows the bit accuracy over the generated frames. From left to right, we present the results of I2V generation from WaTeRFlow-watermarked images while varying \texttt{motion\_bucket\_id}, followed by the results obtained by feeding images watermarked by WaTeRFlow and the baselines into CogVideoX. Finally, we show the results of generating frames with Vidu using images watermarked by WaTeRFlow and the baselines.}
\label{fig:all_supfigs}
\vspace{-5pt}
\end{figure*}

\begin{figure*}
\centering
\includegraphics[width=\linewidth]{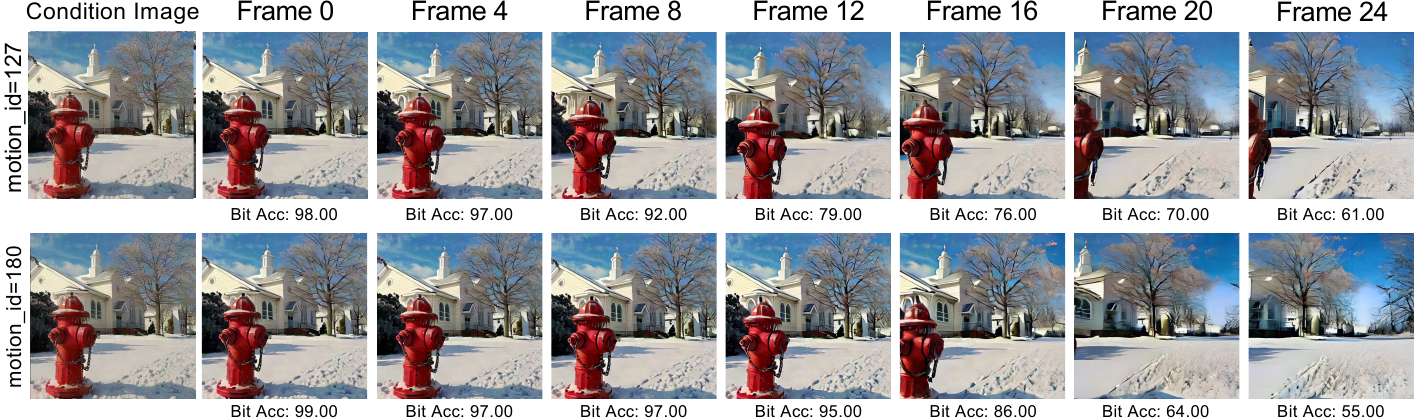}
\vspace{-19pt}
% \caption{\textbf{Qualitative results with respect to \texttt{motion\_bucket\_id}.} The first and second rows show frames generated from watermarked images using our proposed WaTeRFlow with different motion\_bucket\_id values. Top: Frames generated with $\text{motion\_bucket\_id}=127$. Bottom: Frames generated with $\text{motion_bucket_id}=180$, in which the fire hydrant has completely disappeared in the last frame.}
\caption{\textbf{Qualitative results with respect to \texttt{motion\_bucket\_id}.}
The first and second rows show frames generated from watermarked images using our proposed WaTeRFlow with different motion\_bucket\_id values. Top: Frames generated with $\text{motion\_bucket\_id}=127$. Bottom: Frames generated with $\text{motion\_bucket\_id}=180$, in which the fire hydrant has completely disappeared in the last frame.}
\label{supfig:motionid}
\vspace{-10pt}
\end{figure*}

In this section, we experimentally analyze how the bit accuracy changes with the degree of motion in videos generated via image-to-video (I2V) from watermarked images. We use the XT variant of Stable Video Diffusion (SVD)~\cite{stablevideodiffusion} as the video generation model, and vary the hyperparameter \texttt{motion\_bucket\_id}, which determines the degree of dynamics in the video, from its default value of $127$ to $180$ when generating videos from watermarked images. Larger values of this parameter lead to videos with more dynamic motion. Fig.~\ref{supfig:fig1} shows that, as \texttt{motion\_bucket\_id} increases, the overall height of the graph gradually decreases, which corresponds to the results of our proposed method, WaTeRFlow. Furthermore, Fig.~\ref{supfig:motionid} provides qualitative examples illustrating how dynamic the generated videos are when \texttt{motion\_bucket\_id} is set to the default value of $127$ and to $180$, respectively. In both cases, the frames are generated from watermarked images using WaTeRFlow.
% \begin{figure}[ht]
% \vspace{-5pt}
%   \centering
%   \includegraphics[width=\columnwidth]{figure/figure7_00.png}
%   \vspace{-20pt}
%   \caption{\textbf{Empty Caption.}}
%   \label{supfig:fig1}
% \vspace{-10pt}
% \end{figure}

\subsection{Bit Accuracy with Long Video}
In Fig.~4c of the main paper, we present the bit accuracy for frames generated by CogVideoX~\cite{cogvideox} as a plot. This result is obtained using only the first 25 frames out of the 49 frames generated by CogVideoX, so that it can be easily compared with the bit accuracy curve of SVD-XT shown in Fig.~4a of the main paper. Meanwhile, Fig.~\ref{supfig:fig2} shows, for WaTeRFlow and the baselines, the bit accuracy at the frames with even indices among all 49 generated frames. These results demonstrate that WaTeRFlow consistently achieves higher bit accuracy than the baselines over the entire range of frames, including those after the 25th frame.
% \begin{figure}[ht]
% \vspace{-5pt}
%   \centering
%   \includegraphics[width=\columnwidth]{figure/figure8_00.png}
%   \vspace{-20pt}
%   \caption{\textbf{Empty Caption.}}
%   \label{supfig:fig2}
% \vspace{-10pt}
% \end{figure}

\subsection{Commercial Video Generation Service}
In Fig.~4 of the main paper, we evaluated the robustness of each watermarking method by measuring bit accuracy using U-Net based SVD-XT~\cite{stablevideodiffusion} and Diffusion Transformer (DiT) based CogVideoX~\cite{cogvideox}. In this section, instead of open-source video generation models such as SVD-XT and CogVideoX, for which the papers and code are publicly available, we present results obtained with a commercial, closed-source AI video generation platform. Specifically, we conduct experiments using the Vidu Q2 model provided by Vidu~\cite{vidu}. Fig.~\ref{supfig:fig3} shows the qualitative evaluation results for the frames generated by Vidu. The 100 original images used in this experiment are identical to those used to produce the plots in Fig.~4 of the main paper. Fig.~\ref{supfig:fig3} shows that our proposed WaTeRFlow achieves the highest bit accuracy in the generated frames compared to baselines, even on a real commercial video generation platform.
% \begin{figure}[ht]
% \vspace{-5pt}
%   \centering
%   \includegraphics[width=\columnwidth]{figure/figure9_00.png}
%   \vspace{-20pt}
%   \caption{\textbf{Empty Caption.}}
%   \label{supfig:fig3}
% \vspace{-10pt}
% \end{figure}

\subsection{Analysis of Watermarked Image Quality}
\begin{figure}[h]
  \centering
  \includegraphics[width=\columnwidth]{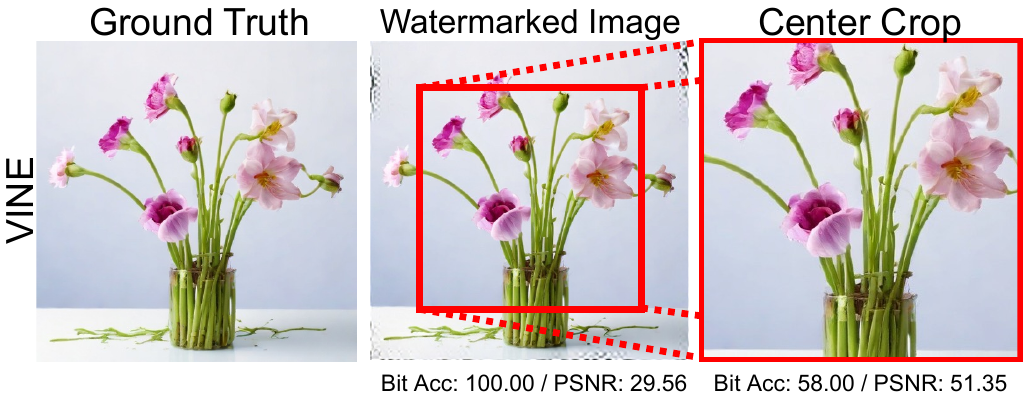}
  \vspace{-10pt}
  \caption{\textbf{Qualitative results of VINE.} This figure visually illustrates why images watermarked with VINE~\cite{vine} yield low PSNR. From left to right, we show the original image, the watermarked image, and the image obtained by applying a center crop that retains 70\% of the height and width. Below each image, we report the corresponding bit accuracy and PSNR.}
  \label{supfig:vine}
\vspace{-5pt}
\end{figure}

In the second None column of Tab.~1 in the main paper, we report the average bit accuracy over video frames generated by SVD-XT~\cite{stablevideodiffusion} for each image watermarking method. In this setting, excluding our proposed WaTeRFlow, VINE~\cite{vine} achieves the highest average bit accuracy. However, in Tab.~2 of the main paper, VINE exhibits a relatively low PSNR. Figure~\ref{supfig:vine} explains this by showing that embedding a watermark with VINE introduces strong noise along the image boundary, which in turn results in low PSNR. If this boundary region is treated as removable and a center crop is applied, PSNR increases significantly, but the watermark then becomes almost unrecoverable. This boundary-focused watermark signal also causes visible noise around frame borders in video frames generated from watermarked images, as seen in Fig.~3 in the main paper. In addition, further experiments with center and inverse center crops, reported in Tab.~\ref{tab:bitacc_crop}, show that baseline methods tend to hide watermark signals primarily near the center or the boundary of the image.
\section{Additional Ablation Study}
\label{additionalablationstudy}
\subsection{Semantic Preservation Loss}
We conducted an ablation study related to the semantic preservation loss in Sec.~5.3 of the main paper. In this section, we present a more detailed analysis of how the bit accuracy of the first generated frame varies with the value of $\lambda_{\text{sem}}$. We ablate the weight $\lambda_{\text{sem}}$ of the semantic preservation loss in Eq.~9 of the main paper. As shown in Table~\ref{tab:lambda_clip}, disabling this term, $\lambda_{\text{sem}} = 0$, reduces the first frame bit accuracy after I2V to $89.63\%$, whereas any non-zero weight in the range $[5\times10^{-5}, 10^{-3}]$ consistently yields $94.82$--$96.93\%$ accuracy, i.e., an absolute gain of about $5$--$7$ percentage points. This confirms that preserving the semantics between the original and watermarked images is important for retaining the embedded watermark in the first frame generated by the I2V model.

\section{Further Experimental Details}
\label{furtherexperimentaldetails}
\subsection{CogVideoX Text Prompts}
This section describes the text prompts used when generating videos with CogVideoX~\cite{cogvideox}. Tab.~\ref{tab:prompts} shows the text prompts used in the CogVideoX experiments throughout the main paper and the supplementary material. In particular, they correspond to the prompts associated with the 100 images used in the experiment shown in Fig.~4 of the main paper. The listed text prompts were applied identically to both our method and the baseline when generating videos from watermarked images using CogVideoX.

\subsection{Resolution Scaling}
TrustMark~\cite{trustmark} and VINE~\cite{vine} propose, as shown in Algorithm~\ref{alg:resolution_scaling}, a method to adapt any watermarking model so that it can handle arbitrary image resolutions. This approach is designed to preserve the visual quality of watermarked images while maintaining the model’s robustness to image transformations it can inherently handle at the native resolution used during training. In all experiments reported in the main paper and the Supplementary Material, we apply this resolution-scaling procedure uniformly to all image watermarking methods so that they can be fairly evaluated at a common resolution of $512 \times 512$.

\section{Failure Case}
\label{failurecase}
\begin{figure*}
\centering
\includegraphics[width=\linewidth]{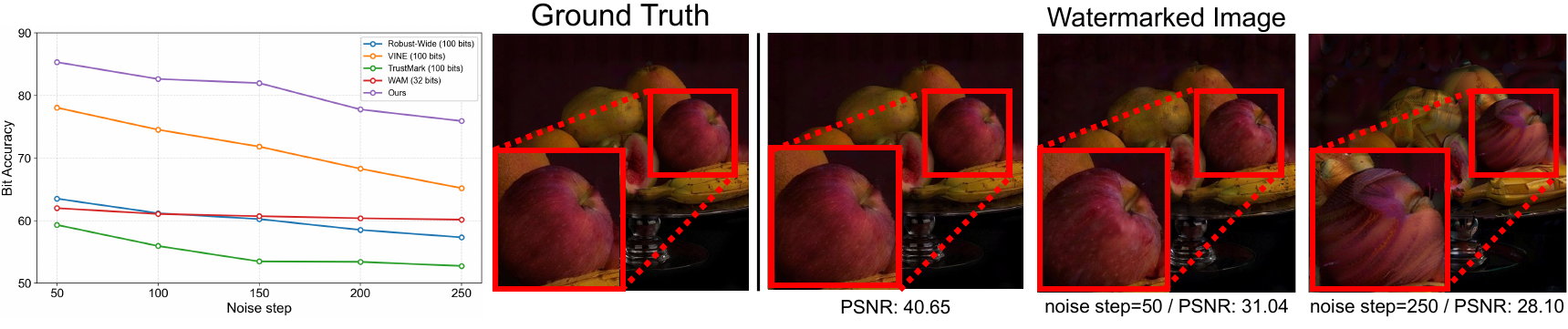}
\vspace{-19pt}
\caption{\textbf{Quantitative and Qualitative Results of Image Regeneration.}  From left to right, we first plot the \textit{average bit accuracy over frames} produced by the I2V generation after stochastically regenerating the  watermarked image with an increasing number of noise steps.  Next, we sequentially show the original image and the watermarked image generated by our proposed method, WaTeRFlow. For the watermarked image, we visualize two cases, one with, regeneration using a relatively low number of noise steps and the other with regeneration using a relatively  high number of noise steps.}
\label{supfig:regen}
\vspace{-10pt}
\end{figure*}

We consider image regeneration as a representative attack in which an adversary attempts to remove embedded image watermarks. In this process, a given image is mapped to a noisy intermediate state along the diffusion trajectory and then denoised back to the image space by a trained diffusion model, producing an output that is visually similar but resampled at the pixel level. Because this procedure weakens fine textures and watermark signals, diffusion-based regeneration constitutes a strong watermark removal attack. In Tab.~1 of the main paper, we regenerate watermarked images using a VP noise schedule~\cite{diffusion_model} and then measure the average bit accuracy in the subsequent image-to-video generation. The graph in Fig.~\ref{supfig:regen} shows the average bit accuracy of our method and the baselines over frames generated by SVD-XT~\cite{stablevideodiffusion} under a stochastic regeneration setting, as we vary the noise step. Our WaTeRFlow achieves the highest average bit accuracy across all evaluated noise steps, though its accuracy falls below 80\% once the noise step exceeds $200$. As shown by the qualitative results in Fig.~\ref{supfig:regen}, such attacks using large noise can suppress watermarks but severely degrade the visual quality of the original image, which limits their practicality as realistic attack scenarios.
\begin{table}[t]
\centering
\resizebox{\columnwidth}{!}{%
\begin{tabular}{ccccc}
\toprule
\multirow{3}{*}{Method} 
& \multirow{3}{*}{\begin{tabular}{@{}c@{}}Cap.\\(Bits)\end{tabular}}
& \multicolumn{3}{c}{Watermarked Image} \\
\cmidrule(lr){3-5}
& & $|\text{Center} - \text{Inverse}|$ (\%) $\downarrow$
  & \multicolumn{2}{c}{Bit Acc. (\%) $\uparrow$} \\
\cmidrule(lr){4-5}
& & & Center Crop (70\%) & Inverse Center Crop (70\%) \\ 
\midrule
\multicolumn{1}{l}{Robust-Wide~\cite{robustwide}} & 100  & \underline{3.26} & 82.33 & 79.07 \\
\multicolumn{1}{l}{VINE~\cite{vine}}              & 100  & 43.27           & 56.73             & 99.99 \\
\multicolumn{1}{l}{TrustMark~\cite{trustmark}}    & 100  & 29.57           & 97.98    & 68.41 \\
\multicolumn{1}{l}{WAM~\cite{wam}}                & 32   & 20.54           & 75.75             & 96.29 \\
\multicolumn{1}{l}{WaTeRFlow (Ours)}              & 100  & \textbf{1.35}   & 78.73             & 77.38 \\
\bottomrule
\end{tabular}%
}
\caption{\textbf{Comparison of center and inverse center cropping on watermarked images.}
We report bit accuracy under center crop and inverse center crop on watermarked images and show the difference between the two cases. These results indicate that existing baseline methods tend to embed watermark signals primarily in the central or peripheral regions of the image.}
\label{tab:bitacc_crop}
\vspace{-15pt}
\end{table}

\section{Additional Qualitative Results}
\label{additionalqualitativeresults}
In this section, we provide additional qualitative evaluations beyond those presented in the main paper.
\subsection{Heatmap}
\begin{figure*}
\centering
\includegraphics[width=\linewidth]{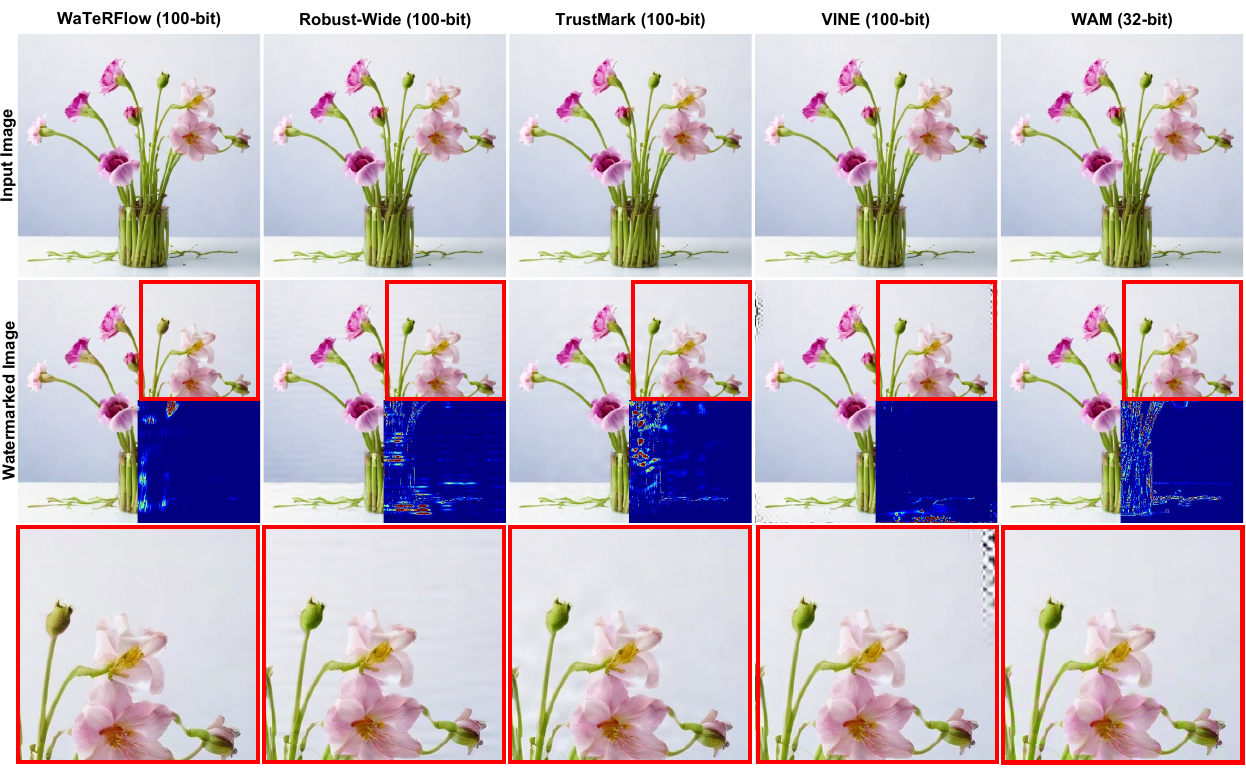}
\vspace{-19pt}
\caption{\textbf{Watermarked Image and Heatmap.} Along with the original image, we sequentially present the resulting images obtained by inserting watermarks using WaTeRFlow and the baselines, as well as the corresponding heatmaps that visualize the regions altered by the watermark as if seen through an X-ray. Top: Original image. Middle: Watermarked images generated by each method and their corresponding heatmaps. Bottom: Zoomed-in crops of the watermarked images.}
\label{supfig:diffmap}
\vspace{-15pt}
\end{figure*}

In this section, we present the original images, the watermarked images, and the heatmaps between them. The heatmaps are generated as follows. First, we compare the original and watermarked images pixel by pixel and combine the changes in the RGB channel values at each location into a single scalar value to calculate a grayscale difference. Next, based on the distribution of these difference values, we rescale them to the range \([0, 1]\), such that values with almost no change fall at the lower end of the range and very large changes move to the upper end. Finally, we apply a colormap to this normalized heatmap, rendering regions with small changes in blue and regions with large changes in red, thereby enabling an intuitive visualization of areas that are strongly affected by watermark insertion. The corresponding results can be found in Fig.~\ref{supfig:diffmap}.
\begin{algorithm}[t]
\small
\caption{Resolution scaling}
\label{alg:resolution_scaling}
\KwInput{Input image $I$, binary watermark $w$}
\KwOutput{Watermarked image $I_w$}
\KwModel{Encoder $E$ trained on the resolution of $U \times V$}
$H, W \leftarrow \mathrm{Size}(I)$\\
$I \leftarrow I / 127.5 - 1$\\
$I' \leftarrow \mathrm{interpolate}(I, (U, V))$\\
$\text{res}' \leftarrow E(I') - I'$\\
$\text{res} \leftarrow \mathrm{interpolate}(\text{res}', (H, W))$\\
$I_w \leftarrow \mathrm{clamp}(I + \text{res}, -1, 1)$\\
$I_w \leftarrow I_w \times 127.5 + 127.5$
\end{algorithm}
\begin{table}[t]
    \centering
    \footnotesize
    \begin{tabular*}{\columnwidth}{@{\extracolsep{\fill}}l|cccc}
        \toprule
        \multicolumn{1}{c|}{$\lambda_{\text{sem}}$} & Bit Acc. (\%) & PSNR $\uparrow$ & SSIM $\uparrow$ & LPIPS $\downarrow$ \\
        & First frame & & & \\
        \midrule
        $10^{-3}$        & 96.93 & 38.83 & 0.9902 & 0.0291 \\
        $5\times10^{-4}$ & 96.22 & 38.18 & 0.9873 & 0.0351 \\
        $10^{-4}$        & 95.99 & 38.19 & 0.9873 & 0.0358 \\
        $5\times10^{-5}$ & 94.82 & 38.39 & 0.9906 & 0.0310 \\
        $0$ (no $\mathcal{L}_{\text{sem}}$) & 89.63 & 40.44 & 0.9920 & 0.0299 \\
        \bottomrule
    \end{tabular*}
    \caption{\textbf{Ablation on semantic preservation loss.} 
$\lambda_{\text{sem}}$ denotes the weight of the semantic preservation loss introduced in Eq.~9 of the main paper. 
PSNR, SSIM, and LPIPS measure the quality of the watermarked images produced by WaTeRFlow. The row no $\mathcal{L}_{\text{sem}}$ indicates the setting where all other contributions of our method are kept, but the semantic preservation loss term is removed.}
    \label{tab:lambda_clip}
    \vspace{-10pt}
\end{table}

\subsection{Additional I2V Qualitative Results}
% In this section, we further present qualitative results of videos generated by SVD-XT and CogVideoX from watermarked input images using the proposed WaTeRFlow and baseline methods. Fig.~\ref{supfig:svd-xt} shows the results generated by SVD-XT, and Fig.~\ref{supfig:cogvideox} shows those generated by CogVideoX.
For additional qualitative results on videos generated from watermarked images using SVD-XT~\cite{stablevideodiffusion} and CogVideoX~\cite{cogvideox}, please refer to the project page provided together with the Supplementary Material.

\clearpage

\clearpage
\onecolumn
{\small
\begin{longtable}{r p{0.8\textwidth}}
\toprule
Index & Prompt \\
\midrule
\endfirsthead
\toprule
Index & Prompt \\
\midrule
\endhead
\bottomrule
\endfoot
0 & The image depicts a charming scene featuring three teddy bears interacting with a small birdhouse in an outdoor setting. \\
1 & The image shows a section of railway tracks with a focus on the foreground where the tracks are clearly visible, showing signs of wear and tear such as rust and discoloration. \\
2 & The image shows a dog dressed in a black leather jacket with a white collar, sitting on the seat of a red motorcycle. \\
3 & The image shows two plush teddy bears sitting side by side against a plain white background. \\
4 & The image shows a neatly made bed with a patchwork quilt featuring various patterns in shades of yellow, gray, and white. \\
5 & The image shows two sandwiches placed inside an oven. \\
6 & The image depicts an urban street scene with a focus on a stop sign in the foreground. \\
7 & The image appears to be the cover of a National Geographic magazine. \\
8 & The image shows two sandwiches being toasted in an oven. \\
9 & The image depicts a serene marina scene with a prominent statue of a giraffe in the foreground. \\
10 & The image depicts a man sitting in a modern office environment, holding a black tablet device in his hands. \\
11 & The image depicts the interior of a modern train carriage. \\
12 & The image depicts a young woman lying on her side on a bed, her head resting on a pillow and her body partially covered by a blue sheet or blanket. \\
13 & The image depicts an indoor equestrian arena where several riders on horseback are engaged in activities, likely practicing or participating in a competition. \\
14 & The image depicts a cozy living room decorated for the Christmas season. \\
15 & The image shows a man with short, dark hair, dressed in a formal blue suit jacket and light blue dress shirt, sitting comfortably in a modern office environment. \\
16 & The image shows a well-worn plush teddy bear sitting upright on a wooden surface in an antique shop or museum setting. \\
17 & The image depicts a steaming cup of coffee placed on a saucer. \\
18 & The image depicts a classroom setting with several students seated at desks. \\
19 & The image depicts a young woman standing in a well-lit kitchen, placing or removing a baking dish from a stainless steel oven. \\
20 & The image shows a medium-sized brown dog, possibly a mixed breed, standing on a windowsill. \\
21 & The image shows a plush brown teddy bear sitting on a light-colored surface in front of a wall that has a height measurement chart painted on it. \\
22 & The image depicts a classroom setting with several students seated in rows, concentrating on writing or taking an exam. \\
23 & The image depicts a classroom setting with several students seated at desks, each working on assignments or taking notes. \\
24 & The image depicts a construction worker wearing high-visibility clothing operating a hydraulic excavator on a construction site. \\
25 & The image shows a busy classroom environment where multiple students are seated at desks, working intently on tasks or exams. \\
26 & The image shows a young girl with long blonde hair, dressed in a white top and blue jeans, standing on a grassy lawn in a park or backyard setting. \\
27 & The image depicts a group of school-aged children seated around a rectangular table in a classroom setting. \\
28 & The image appears to be a scanned or photographed page from a book or academic text written in German. \\
29 & The image depicts a man walking down a city street at night. \\
30 & The image shows the interior of a classroom filled with students seated at desks arranged in rows. \\
31 & The image shows an indoor ice hockey rink where a pickup game is taking place. \\
32 & The image shows a close-up view of a person's torso and arms as they sit at a desk or table, using a pen to write in an open notebook or planner. \\
33 & The image depicts an outdoor poolside scene at what appears to be a residential or private property. \\
34 & The image shows a dog partially outside a window or door, holding a plush toy of a squirrel in its mouth. \\
35 & The image shows a medium-sized light brown dog standing on a windowsill looking outside. \\
36 & The image depicts a close-up of a teddy bear with long, furry brown hair. \\
37 & The image depicts an indoor setting where several individuals are engaged in what appears to be a discussion or meeting. \\
38 & The image appears to be a scanned or printed page from a German-language academic or technical book. \\
39 & The image depicts a maritime scene featuring a large sailing ship docked at a harbor. \\
40 & The image depicts a dog standing on the windowsill of a house, looking outside. \\
41 & The image depicts an indoor scene where a man is sitting in a bar or pub setting. \\
42 & The image shows a light brown teddy bear wearing a red sweater with the word London and the letter E on it, sitting in front of a scenic background featuring the Big Ben clock tower and the Palace of Westminster in London. \\
43 & The image depicts a quaint village street lined with traditional stone cottages featuring chimneys and slate roofs. \\
44 & The image shows a person performing a bicycle stunt in mid-air, likely in a skate park or similar recreational area. \\
45 & The image shows a woman sitting at a table in a bright office or meeting room, engaged in a discussion or presentation. \\
46 & The image shows a close-up of a can of Bunnaberg \& Cola. \\
47 & The image shows a group of young women sitting in what appears to be a classroom or training room, wearing bright green vests and smiling towards the camera. \\
48 & The image depicts a panel discussion or interview setting involving three individuals seated on high chairs or stools in front of a bright yellow backdrop. \\
49 & The image shows two action figures of Teenage Mutant Ninja Turtles, specifically Leonardo and Donatello, riding skateboards. \\
50 & The image depicts a black folder laying open on a table with several documents neatly arranged on both sides. \\
51 & The image shows a young woman in a white lab coat standing at a lab bench in a classroom or laboratory setting. \\
52 & The image shows a group of individuals seated in what appears to be a classroom or training room setting. \\
53 & The image depicts a large yellow truck, likely used for industrial or construction purposes, parked on a paved area near a modern building. \\
54 & The image depicts a dog standing on the ocean shore during sunset. \\
55 & The image shows a football player in action on the field during a game. \\
56 & The image depicts a black and white portrait of a dog, possibly a mixed breed or a spaniel type. \\
57 & The image shows a group of people sitting in a classroom or training room setting. \\
58 & The image shows a stadium filled with a crowd of enthusiastic fans, many of whom are wearing yellow shirts or jerseys and waving yellow flags. \\
59 & The image shows a professional ice hockey game in progress. \\
60 & The image depicts a woman social worker or case manager visiting an elderly man in his home. \\
61 & The image shows two brown teddy bears sitting on a couch or sofa. \\
62 & The image depicts a close-up of a playful puppy standing on a beach or sandy area near a waterfront. \\
63 & The image depicts a close-up of a calculator with a pencil resting on an open notebook or planner. \\
64 & The image depicts a classic London street scene with an iconic red double-decker bus prominently featured in the foreground. \\
65 & The image appears to be a scanned or photographed page from an academic book or textbook in German. \\
66 & The image shows a serene beach scene with two wooden lounge chairs placed on a white sandy shore. \\
67 & The image shows an outdoor swimming pool in a residential backyard or small resort setting. \\
68 & The image shows a busy job interview or career fair setting. \\
69 & The image appears to be a scanned or photocopied page from a German-language book. \\
70 & The image depicts a modern office building with a glass facade located in an urban setting. \\
71 & The image depicts a lively outdoor music concert or festival scene. \\
72 & The image depicts a dog lying down and resting its head on what appears to be the leg or knee of a person. \\
73 & The image shows a dog standing on a windowsill partially obscured by white curtains on either side. \\
74 & The image depicts a panel discussion or interview setup in a studio environment. \\
75 & The image depicts a motorcycle race on a racetrack, featuring two riders leaning into a curve at high speed. \\
76 & The image shows a black and white photograph of a football locker room. \\
77 & The image depicts a man standing in a kitchen or living room area, holding a yellow cleaning bucket in his right hand. \\
78 & The image shows a dog standing on the windowsill of a house or building, looking outside through an open window. \\
79 & The image shows a close-up view of an audio mixing console in a recording studio or live sound setting. \\
80 & The image depicts a physical therapy or rehabilitation session taking place in a medical or clinical setting. \\
81 & The image depicts a mother and her newborn baby in a hospital room shortly after birth. \\
82 & The image shows three green military-style armored trucks parked in a line on a dirt or gravel surface. \\
83 & The image depicts a pair of boots displayed in an artistic arrangement. \\
84 & The image depicts a group of seven people in an office setting, all smiling and giving thumbs-up gestures towards the camera. \\
85 & The image depicts a modern laboratory setting with several individuals engaged in scientific research or experimentation. \\
86 & The image depicts a classic red double-decker bus, reminiscent of those commonly seen in London, England. \\
87 & The image shows the facade of a brick building with a prominent clock mounted on it. \\
88 & The image shows a British Rail Class 47 diesel-electric locomotive at a railway station platform. \\
89 & The image depicts a smartphone with a digital representation of a credit card displayed on its screen. \\
90 & The image captures a skateboarder performing a trick in an urban skate park setting. \\
91 & The image depicts a clock tower situated in an urban environment. \\
92 & The image shows the iconic clock face of Big Ben, a part of the Elizabeth Tower in London, England. \\
93 & The image shows a bathroom with a toilet as the central focus. \\
94 & The image depicts a scene from a theatrical performance or a play. \\
95 & The image depicts a charming urban scene featuring a clock tower that stands prominently in the center. \\
96 & The image shows a white vehicle parked on a street or driveway. \\
97 & The image appears to be a promotional poster for an event or performance. \\
98 & The image depicts an urban street scene at dusk or early evening, characterized by the illuminated windows of tall buildings in the background, suggesting a cityscape with modern architecture. \\
99 & The image features a person standing in front of a red, vertically slatted background. \\
\end{longtable}
\captionsetup{hypcap=false}
\captionof{table}{\textbf{Text prompts used in the CogVideoX experiments.} 
Text prompts corresponding to the 100 images used in the experiment shown in Fig.~4 of the main paper, which are used to generate videos with CogVideoX~\cite{cogvideox} throughout the main paper and the Supplementary Material. The same prompts are applied to both our method and the baselines when generating videos from watermarked images.}
\label{tab:prompts}
}
\twocolumn

%%%%%%%%% REFERENCES
\clearpage
{\small
\bibliography{_suppl,main}
}